\documentclass[10pt,twocolumn,letterpaper]{article}

 \usepackage[pagenumbers]{cvpr} % To force page numbers, e.g. for an arXiv version

\usepackage[dvipsnames]{xcolor}

\definecolor{cvprblue}{rgb}{0.21,0.49,0.74}
\usepackage[pagebackref,breaklinks,colorlinks,citecolor=cvprblue]{hyperref}
\usepackage{appendix}
\usepackage{array,multirow,graphicx}
\usepackage{stmaryrd}
\usepackage{amsmath}
\usepackage{amssymb}
\usepackage[none]{hyphenat}
\usepackage{caption}
\def\paperID{15124} % *** Enter the Paper ID here
\def\confName{CVPR}
\def\confYear{2024}

\title{SG-PGM: Partial Graph Matching Network with Semantic Geometric Fusion\\ for 3D Scene Graph Alignment and Its Downstream Tasks}

\author{Yaxu Xie \quad Alain Pagani \quad Didier Stricker \vspace{0.1cm} \\
German Research Center for Artificial Intelligence\\
{\tt\small firstname.lastname@dfki.de}
}

\begin{document}
\maketitle
\begin{abstract}
Scene graphs have been recently introduced into 3D spatial understanding as a comprehensive representation of the scene.
The alignment between 3D scene graphs is the first step of many downstream tasks such as scene graph aided point cloud registration, mosaicking, overlap checking, and robot navigation. 
In this work, we treat 3D scene graph alignment as a partial graph-matching problem and propose to solve it with a graph neural network. 
We reuse the geometric features learned by a point cloud registration method and associate the clustered point-level geometric features with the node-level semantic feature via our designed feature fusion module. 
Partial matching is enabled by using a learnable method to select the top-k similar node pairs. 
Subsequent downstream tasks such as point cloud registration are achieved by running a pre-trained registration network within the matched regions.
We further propose a point-matching rescoring method, that uses the node-wise alignment of the 3D scene graph to reweight the matching candidates from a pre-trained point cloud registration method. It reduces the false point correspondences estimated especially in low-overlapping cases.
Experiments show that our method improves the alignment accuracy by 10$\sim$20\% in low-overlap and random transformation scenarios and outperforms the existing work in multiple downstream tasks. Our code and models are available \href{https://github.com/dfki-av/sg-pgm.git}{here}. 
\end{abstract}    
\section{Introduction}
\label{sec:intro}

\begin{figure}[tb]
    \centering
    \includegraphics[width=\linewidth]{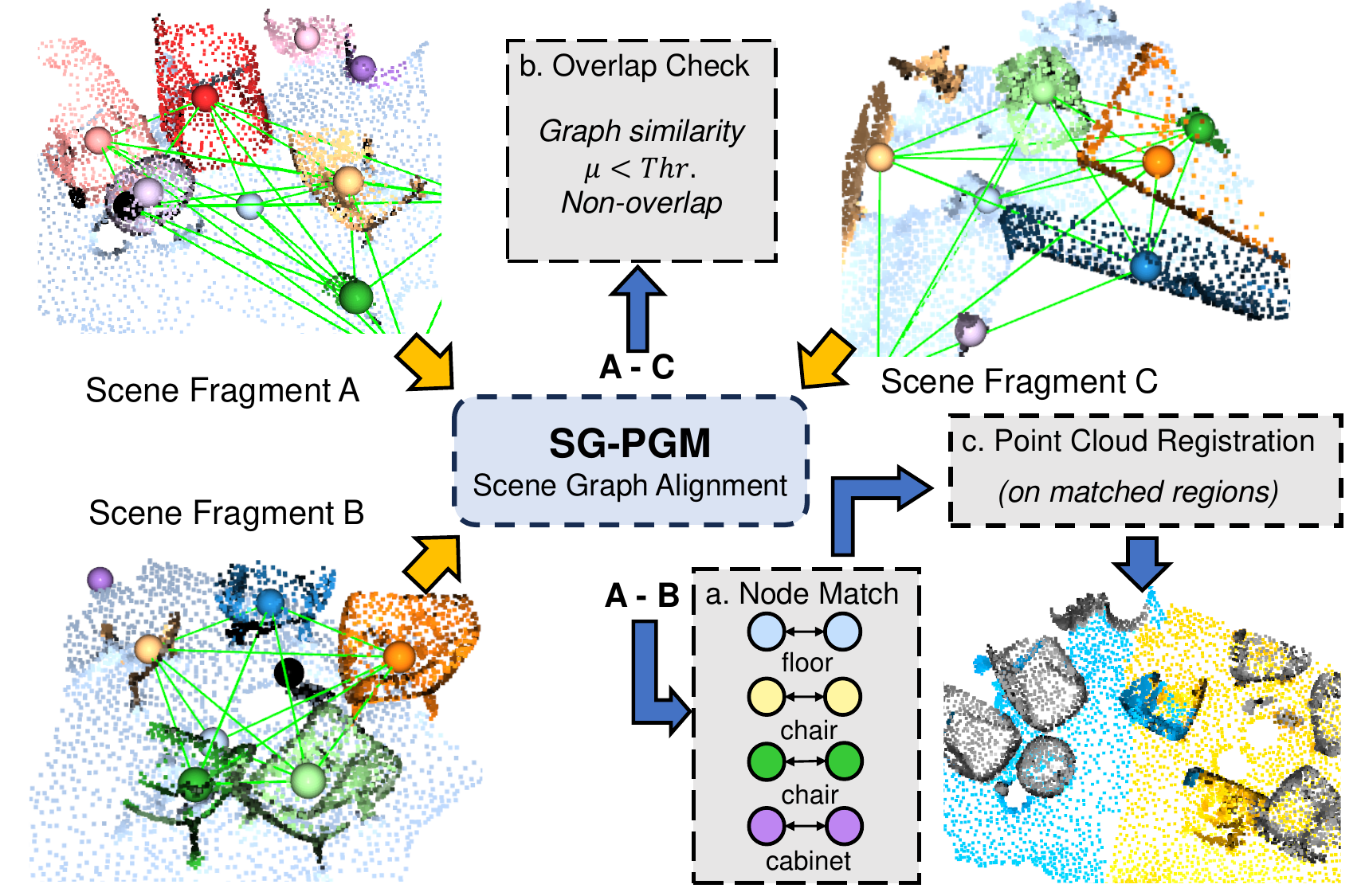}
    \caption{\textbf{SG-PGM}: partial graph matching for 3D scene graph alignment. 
    Semantic and geometric features are fused for object-wise matching between fragments (a), and downstream tasks such as (b) overlap-check and (c) point cloud registration. 
    }
    \label{fig:teaser}
    \vspace{-2mm}
\end{figure}

The 3D semantic scene graph~\cite{armeni20193d,wald2020learning,wu2021scenegraphfusion} is a semantic-rich model for scene representation, which summarizes the scene context in the form of an attributed and directed graph, in which 3D objects and structures as nodes are associated with semantic classes (e.g. sofa, wall), and geometrical semantic relationship between nodes are represented as edges with multiple classes (e.g. \textit{stand\_on}, \textit{supported\_by}).
3D scene graphs support many applications in spatial understanding, such as global localization for SLAM~\cite{gawel2018x, liu2019global,  hughes2022hydra, rosinol2021kimera}, loop-closure detecting~\cite{qian2022towards}, robot navigation~\cite{zhu2021soon, wang2022object}, visual object grounding~\cite{feng2021free}, graph-to-3D manipulation~\cite{graph2scene2021} and augmented reality~\cite{tahara2020retargetable}. 
One of the main problems of the aforementioned applications is searching for the partial alignment of two or more 3D scene graphs. 
As illustrated in Figure~\ref{fig:teaser}, once the alignment between nodes is found, tasks like localization and navigation can be conducted via point cloud registration within the overlapping area. 
Alternatively, the determination of whether scene fragments are overlapped or not can be achieved by analyzing the similarity of 3D scene graphs.

SGAligner~\cite{sarkar2023sgaligner} is the first work specifically focusing on this problem. 
In this work, Sarkar et al.~proposed a neural network that learns a joint multi-modal embedding encoded with semantic, geometric, and structural information for each node entity in the graph, which is trained with cross-modal contrastive loss and outputs the similarity between source and reference graph nodes as the alignment result. 
After the node(object)-level alignment is found, downstream tasks such as point cloud registration are conducted by using a pre-trained registration method, to search point matching within two aligned objects. 
Later, the point pairs of all aligned objects are fed into a graph-cut RANSAC algorithm~\cite{GCRansac2018} to estimate the transformation between the source and reference point clouds. 
Such decoupled design allows SGAligner to be easily plugged into most of the feature-based registration methods.

However, the simplicity of such a two-stage approach comes with some drawbacks: 
First, SGAligner employs PointNet~\cite{qi2017pointnet} to encode object-level geometric embedding. For downstream tasks like point cloud registration or mosaicking, the geometric feature will be extracted twice, once for scene graph alignment and once for registration. 
We find that reusing the geometric feature extracted from more powerful 3D points encoder like Edge Conv~\cite{wang2019dynamic}, FCGF~\cite{choy2019fully} and KP-Conv~\cite{thomas2019kpconv} is more efficient since they are already integrated into recent registration method~\cite{fey2019deep, predator,yu2021cofinet,fu2021robust,qin2022geometric}.
Second, SGAligner achieves a computation complexity less than $\textit{O}\left( N^{2}\right)$ (looping through all possible node pairs) by running registration only on the predicted alignment pairs. However, we argue the complexity can be further reduced by introducing explicit mechanisms, that enable one-to-one matching or even partial alignment to surpass false-positive prediction. 

Addressing the aforementioned aspects, we first define the 3D scene graph alignment as a \textbf{partial graph matching problem}.
We build our graph matching neural network following the linear assignment formalism: encoding edge information into node features with the graph convolution and searching node-to-node matching via the Sinkhorn decoder~\cite{sinkhorn1964relationship}.
We reuse the point features learned by the backbone of the point cloud registration method and cluster the point-wise geometric feature into entity nodes via our designed Point to Scene Graph Fusion module (P2SG). 
We additionally enable explicit partial matching by employing differentiable top-k method~\cite{wang2023deep} to select the $k$ most likely matching pairs. This further increases the alignment accuracy and reduces the false-positive prediction.

Moreover, we design a Superpoint Matching Rescoring method using the predicted scene graph node alignment as the semantic level prior to guiding the point correspondence estimation during registration. 
We further reduce the search space of point-to-point matching during registration by masking out the non-aligned objects of both scene fragments. 
We conduct point cloud registration only once between the predicted overlap regions of scene fragments, instead of traversing through node pairs as done in~\cite{sarkar2023sgaligner}. By employing this strategy, we reduce the inference time, while retaining the long-distance cross-object geometric feature potentially encoded by registration methods~\cite{qin2022geometric, yu2023peal}.

We showcase the effectiveness of our approach, by experimenting with scene graph alignment and its downstream tasks: overlap-checking, point cloud registration, point cloud mosaicking, and alignment with dynamics on the 3RScan~\cite{wald2019rio,wald2020learning} dataset. 
Results show that our approach significantly improves the alignment accuracy by 10$\sim$20\% compared to~\cite{sarkar2023sgaligner}, especially when transformation $T\neq I_{4}$ exists between scene fragments. It reduces the rotation error by 50\%, and the translation error by 24\% on the point cloud registration task, compared to~\cite{sarkar2023sgaligner} while keeping the registration RANSAC-free. 
We also conduct ablation studies to visually demonstrate the effecting mechanism of our proposed Superpoint Matching Rescoring and compare different strategies of using alignment results on registration. 
We summarize the contributions of this paper as follows:
\begin{enumerate}
    \item A graph neural network (SG-PGM) for partial graph matching to solve 3D scene graph alignment. 
    \item  The Point to Scene Graph Fusion module and the soft top-k method for increasing alignment accuracy.
    \item The Superpoint Matching Rescoring method for guiding the point matching with scene graph alignment results.
    \item Revisiting the strategies to stimulate the potential of using 3D scene graph alignment for downstream tasks.
    
\end{enumerate}

\section{Related Work}
\label{sec:rw}

%In this section, we first explain the prerequisites of our work: approaches to generate 3D semantic scene graphs and existing datasets. Then we give a summary of related literature on two fundamental components in our work: learning-based (sub)graph matching and point cloud registration. Finally, we discuss the existing works for 3D scene graph alignment and its downstream tasks. 

\noindent \textbf{3D Semantic Scene Graph} can be estimated from a video sequence, panoramic image or point could in a bottom-up fashion. Armeni et al.~\cite{armeni20193d} design a semi-automatic framework based on object detector and multi-view consistency and use it to extend the 2D scene graph in~\cite{krishna2017visual} into 3D space. Wald et al.~\cite{wald2020learning} present their 3D scene graph dataset extended from 3RScan~\cite{wald2019rio}, in which object and structure nodes are annotated with multiple geometric relationships as edges. Their proposed network estimates a 3D semantic scene graph from the point cloud of the scene. Later Wu et al. proposed an incremental method to predict 3D scene graphs from RGB-D~\cite{wu2021scenegraphfusion} and RGB~\cite{wu2023incremental} sequence as input. Zhang et al.~\cite{zhang2021knowledge} introduced knowledge learning and knowledge intervention-aided scene graph prediction.

\begin{figure*}[htb]
     \centering
     \begin{subfigure}[b]{0.522\linewidth}
         \includegraphics[width=\textwidth]{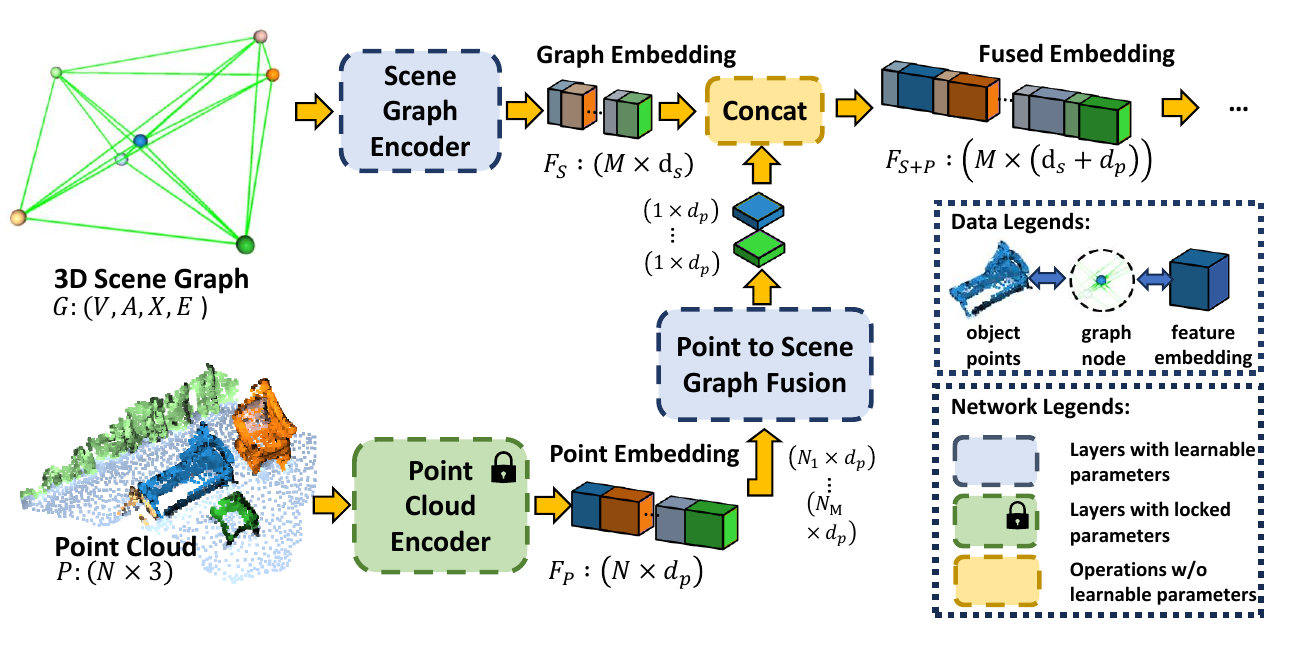}
         \caption{Scene Graph and Point Encoding}
         \label{fig:embed}
     \end{subfigure}
     \begin{subfigure}[b]{0.472\linewidth}
         \includegraphics[width=\textwidth]{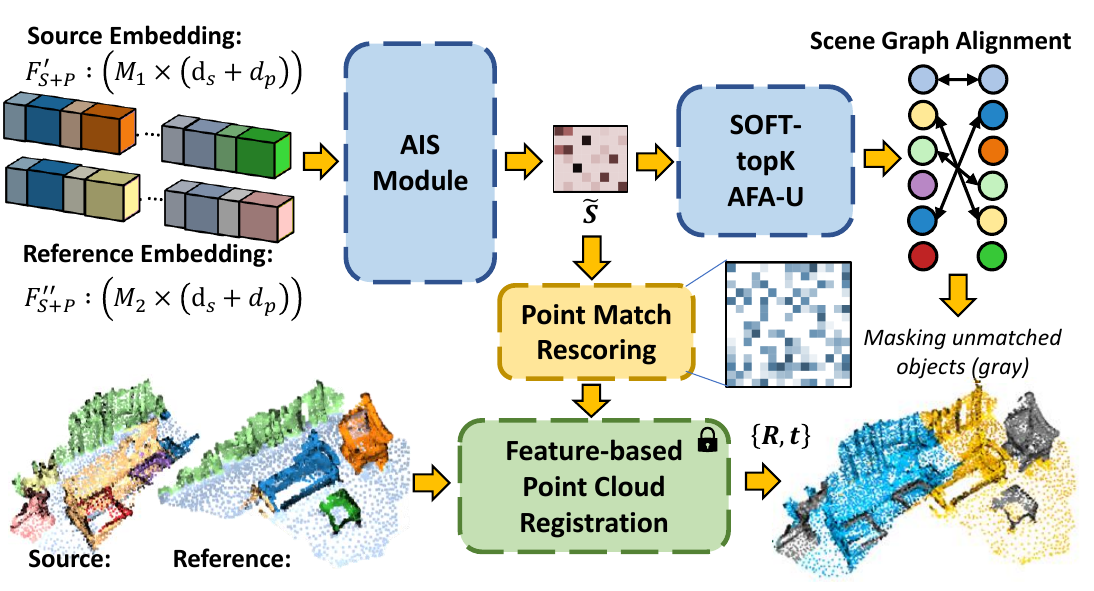}
    \caption{Scene Graph Alignment and Point Cloud Registration}
    \label{fig:match}
     \end{subfigure}
     \caption{\textbf{The network overview} of the proposed system. (a) shows the feature extraction and our proposed \textbf{Point to Scene Graph Feature Fusion} of one single point cloud and its associated 3D scene graph. (b) shows the alignment stage between the source and the reference scene graphs and the registration stage of point clouds with the guidance of our proposed \textbf{Superpoint Matching Rescoring} method. We reuse the pretrained point cloud encoder of the point cloud registration method. Its weights are locked during training.} 
    \label{fig:net} \vspace{-3.5mm}
\end{figure*}

\noindent \textbf{Graph Matching and Subgraph Matching}
share the same goal of finding the one-to-one alignment between graphs, while the latter is also required to determine the existence of a subgraph isomorphism. NeuroMatch~\cite{lou2020neural} presents the first subgraph matching network that estimates the subgraph relationship with the learned order embedding~\cite{mcfee2009partial}. Later works~\cite{lan2021sub, roy2022interpretable} estimate the node or edge correspondence between query and target graphs directly, which makes them similar to many general graph matching works~\cite{fey2019deep, liu2021stochastic, liu2022deep}. 
Graph matching is used in many domains of computer vision, such as object key point detection~\cite{wang2019learning, Zanfir_2018_CVPR} and tracking~\cite{He_2021_CVPR}, SfM and SLAM~\cite{sarlin20superglue}. 
For 3D scene graph alignment, two graphs are usually partially matched.
Wang et al.~\cite{wang2023deep} enable partial graph matching with a differentiable top-k framework to select the most likely matched pairs from the primary one-to-one matching.

\noindent \textbf{Learning-based Point Cloud Registration} can be divided into the end-to-end methods and the feature-based methods.
PointNetLK~\cite{yaoki2019pointnetlk} proposes aligning the global descriptors iteratively via Lucas \&Kanade algorithm~\cite{lucas1981iterative}.
OMNet~\cite{xu2021omnet} introduces overlapping mask prediction into the end-to-end method and enables partial registration. 
The challenge of low overlap scene-level registration is tackled by Huang et al.~\cite{predator}. Their proposed Overlap Attention Module extracts co-contextual features between point clouds and predicts overlapping and match-ability scores during early information exchange. 
Qin et al. propose GeoTransformer~\cite{qin2022geometric} that learns transformation-invariant geometric representation on the level of super-point using Transformer~\cite{vaswani2017attention} with their proposed Geometric Structure Embedding. The correspondence is searched first at the super-point level and then at the point level. A Local-to-Global scheme is designed to solve the transformation with weighted SVD~\cite{besl1992method}.

\noindent \textbf{3D Semantic Graph Alignment and Downstream Tasks} are common in the robotics domain, e.g. using semantic information in the scene to improve the localization accuracy and robustness. 
X-View~\cite{gawel2018x} presents semantic topological graph with nodes assigned with semantic labels and center locations, connected with non-directed edges for global localization. The graph matching is solved by computing the similarity between the random walk descriptors of nodes. Qiao et al.~\cite{qiao2022objects} proposed the Object Relation Graph feature that encodes the deep visual and relationship representations of detected objects. After the more unified form~\cite{wald2020learning} of 3D scene graph was defined recently, Sarkar et al.~\cite{sarkar2023sgaligner} explored in SGAligner 3D scene graph alignment and its downstream applications such as point cloud registration with non-overlap early stopping, point cloud mosaicking, 3D scene alignment with changes. This work proposes the first method for aligning pairs of 3D scene graphs and provides data generation pipelines and benchmarks for each task.

\section{Approach}
\label{sec:app}

\subsection{Scene Graph Matching Network}
\noindent \textbf{Problem Definition.} A 3D scene graph is a graph model with semantic node and edge attributes: $\mathcal{G} = (\mathcal{V}, \textit{\textbf{A}}, \textit{\textbf{X}}, \textit{\textbf{E}})$. 
It consists of a finite set of object nodes $\mathcal{V}=\left \{v_{1},v_{2}, ... , v_{M}\right \}$, an adjacency matrix $A\in \left \{ 0, 1\right \}^{M \times M}$, a node feature matrix $\textit{\textbf{X}} \in \mathbb{R}^{ M\times \cdot}$ and a edge feature matrix $\textit{\textbf{E}} \in \mathbb{R}^{M  \times M \times \cdot}$. 
Additionally, each 3D points of the corresponded point cloud $P = \left \{ \textbf{p}_{i} \in \mathbb{R}^{3} \mid  i = 1, ..., N\right\}$ is assigned to one specific object node with point-to-object map $O: \{1, 2, \ldots, N\} \rightarrow \{1, 2, \ldots, M\}$.

The 3D scene graph may contain noise due to the imperfect output of graph estimation method~\cite{wald2020learning, wu2021scenegraphfusion, zhang2021knowledge, wu2023incremental} and the dynamical scene changes in long-term~\cite{wald2019rio}.
Instead of posing the problem as a graph isomorphism search, we formulate the inexact graph matching as optimizing the following objective function:

\begin{equation}
    \underset{\textbf{S}}{\arg\max} \: f(\textbf{S};\mathcal{G}_{src}, \mathcal{G}_{ref}),
\end{equation}

\noindent in which $\textbf{S} \in \left \{ 0,1\right \}^{M_{src}  \times M_{ref} }$ is the binary permutation matrix that maps nodes between the source graph $\mathcal{G}_{src}$ and the reference graph $\mathcal{G}_{ref}$. We follow~\cite{fey2019deep,kriege2019computing,liu2021stochastic} to further relax the constraint from the Quadratic Assignment Problem to the Linear Assignment Problem, and define the objective function $f\left(\cdot \right)$ as the negative cross entropy between the ground truth \textbf{S} and the approximate matching $\tilde{\textbf{S}}$, which is learned by our neural network $\tilde{\textbf{S}}=nn\left(\mathcal{G}_{src}, \mathcal{G}_{ref} \right)$.

\begin{figure}[htb]
    \centering
    \includegraphics[width=\linewidth]{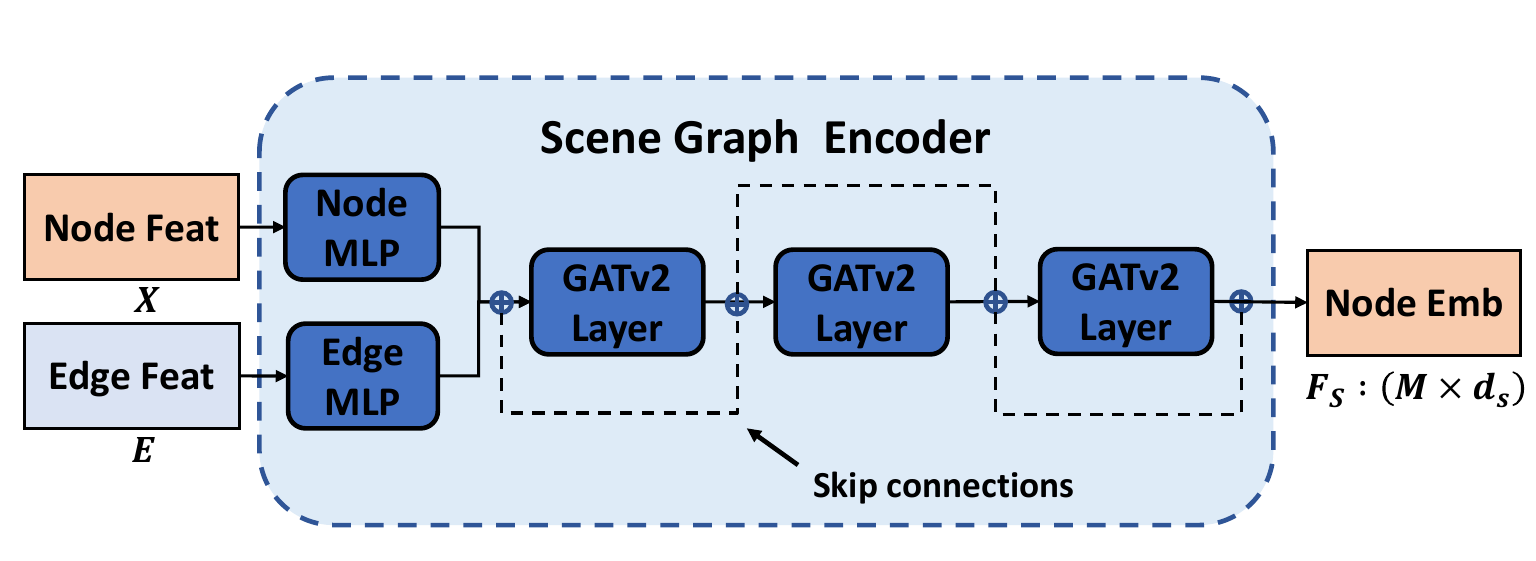}
    \caption{\textbf{Scene graph encoder} with GATv2 layers and learnable skip connections.}
    \label{fig:matchnet}
\end{figure}

\noindent \textbf{Partial Graph Matching Network.} As illustrated in~\ref{fig:embed}, our matching network first projects the semantic node features $\textit{\textbf{X}}$ and semantic edge features $\textit{\textbf{E}}$ of the source and reference graphs into the graph embedding ${F}_{S}$. We then combine the geometric embedding ${F}_{P}$ from the point cloud encoder to form the fused embedding  ${F}_{S+P}$.
In more details, $\textit{\textbf{X}}$ and  $\textit{\textbf{E}}$ are first encoded into
the same dimension $d$ with MLPs, then $n$-layers GATv2~\cite{brody2022how} extract the semantic and topological information of each node in the graph. We built learnable skip connections between layers in the same manner as in~\cite{lou2020neural}, which is theoretically proved in~\cite{xu2021optimization} to converge more efficiently. Thus, the scene graph encoder outputs multi-layers node embedding $ {F}_{S} \in \mathbb{R}^{ M \times d_{s}}$ with $d_{s} = d(n+1)$, as shown in Figure~\ref{fig:matchnet}.

In the alignment and registration stage (shown in Figure~\ref{fig:match}), fused embedding of the source and reference graph is taken by the AIS~\cite{fu2021robust} module to provide a cost matrix that measures the pair-wise similarity.
In this module, the joint scene graph and geometric node embedding ${F}^{ref}_{S+P}$ and ${F}^{src}_{S+P}$ (see Section~\ref{sec:app_p2s}) are used to compute an affinity matrix $\textbf{A}$ by:
\begin{equation}
\label{eqn:ais}
    \textbf{A} = {F}^{ref}_{S+P} \begin{bmatrix} \textbf{W}_{s} & 0 \\ 0 & \textbf{W}_{p} \end{bmatrix} {F}^{src}_{S+P},
\end{equation}
in which $\textbf{W}_{s}$ and $\textbf{W}_{p}$ are the learnable weights for computing the affinity of both node embedding. Then $\textbf{A}$ is normalized via instance normalization and processed by the Sinkhorn~\cite{mena2018learning, sinkhorn1964relationship} operator with an additional row and column of zeros.
This enables nodes without correspondence to be matched to the dummy row and column instead. Now we have the soft matching prediction as a doubly-stochastic matrix $\tilde{\textbf{S}}$, to approximate the one-to-one permutation matrix $\textbf{S}$. 

To explicitly enable partial matching, we employ the pipeline introduced in~\cite{wang2023deep}:  the Soft-topK algorithm first flattens $\tilde{\textbf{S}}$ and selects the $K$ most likely matched candidates, where $K$ is learned by an Attention-fused Aggregation Module~\cite{wang2023deep}, more specifically its AFA-U variant. 
In this module, dummy node features ${F}'_{src}$ and ${F}'_{ref}$ (see App.~\ref{sec:imp_detail}) are formed into a bipartite graph with $\tilde{\textbf{S}}$ as the weighted edges, and are brought to a graph attention layer to predict $\tilde{k} \in \left[ 0, 1\right]$ as a graph similarity score, with $K=\tilde{k}\times\left|M_{ref}\right|$.

\subsection{Point to Scene Graph Feature Fusion}
\label{sec:app_p2s}

If only considering the semantic information in the 3D scene graph, nodes with the same semantic label and the same edge connection to the other nodes are \textbf{symmetric}, e.g. several pillows \textit{lie\_on} a sofa. 
In that case, the subgraph that only consists of these nodes is \textbf{automorphism}. Therefore, their graph embedding $F_{S}$ is identical and it results in unsolvable ambiguity in matching.

Addressing this, we propose to combine the semantic scene graph embedding ${F_{S}}$ with the point geometric embedding ${F_{P}}$ of each object node, in order to form a more distinguishable joint embedding ${F_{S+P}} \in \mathbb{R}^{ M  \times \left ( d_{s} + d_{p}\right )}$.  
Since our scene graph matching network will cooperate with a feature-based point cloud registration network for solving downstream tasks, it is more efficient to share the point-wise geometric feature encoded by the same backbone network than to introduce another point feature encoder for the same aim.

\begin{figure}[htb]
    \centering
    \includegraphics[width=\linewidth]{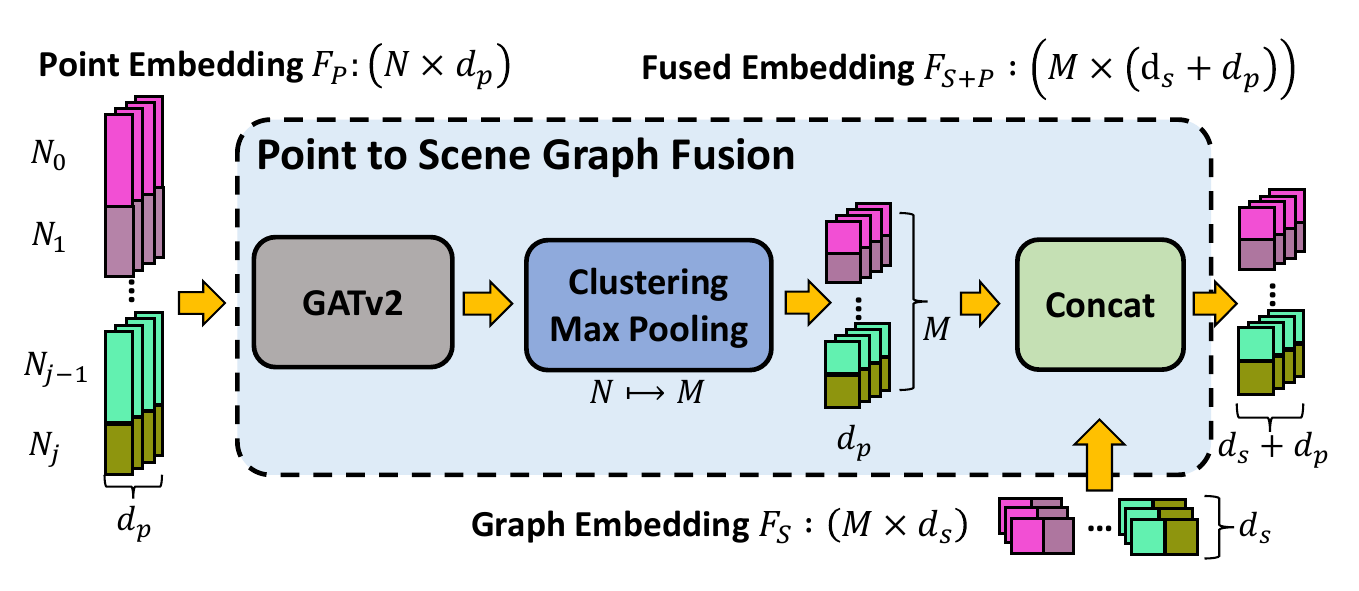}
    \caption{\textbf{P2SG fusion module} projects point-wise geometric features to node-wise geometric embedding and combines it with the semantic scene graph feature. %Due to the sequential-ignorance nature of the aggregation operation in graph neural networks, the order of points has no impact on the result.
    }
    \label{fig:fusion}
\end{figure}
As is illustrated in Figure~\ref{fig:fusion}, we design this novel Point to Scene Graph Fusion module (P2SG) that projects geometric feature $ F_{\mathfrak{p}} \in \mathbb{R}^{  N \times d_{p}}$ of $N$ points to object-level feature $ {F}_{P} \in \mathbb{R}^{M  \times d_{p}}$ of $M$ nodes. 
The module is defined as:

\begin{equation}
\label{eqn:gnn}
\begin{gathered}
        {{F}_{p}}^{'} = f_{\theta} \left( {{F}_{p}} , E_{knn}\right), \\
        {F}_{P} \in \mathbb{R}^{  M  \times d_{p}} \overset{O}{\mapsfrom}  {{F}_{p}}^{'} \in \mathbb{R}^{  N  \times d_{p}},
        \end{gathered}
\end{equation}
where $E_{knn}$ is the k-nearest neighbor edges built according to the Euclidean distance between 3d points, and $f_{\theta}\left(\cdot \right)$ is a GATv2~\cite{brody2022how} layer for aggregating neighbor features. The clustering max pooling operation $\overset{O}{\mapsfrom}$ pools the point-wise geometric feature into the node-wise feature with the point-to-object map $O$.

\subsection{Super-point Matching Rescoring}
Feature-based point cloud registration methods like GeoTransformer~\cite{qin2022geometric} first compare the similarity of points or super-points, to determine the potential point-wise correspondence. Then the transformation can be estimated using weighted SVD~\cite{besl1992method}, RANSAC~\cite{fischler1981random}, or its variant~\cite{GCRansac2018}. However, only computing the geometric similarity between points will potentially cause incorrect matching, if two points have very similar local geometric features but globally not even belonging to the same object.

We propose the Super-point Matching Rescoring method that uses the semantic similarity learned by our scene graph matching network to reweight the point-wise matching score. 
Having the scene graph node matching matrix as $\tilde{\textbf{S}} \in \mathbb{R}^{M \times M}$, the super-point matching matrix\footnote{Please refer Eq.9 in GeoTransfomer~\cite{qin2022geometric} for more detail.} $\textbf{C}$  can be rescored to ${\textbf{C}}'$ with:
\begin{equation}
    {\textbf{C}}' = \textbf{C} + \gamma \textbf{R},
\end{equation}
where $\textbf{R} \in \mathbb{R}^{N \times N}$ is the rescoring matrix expanded from $\tilde{\textbf{S}}$ using the point-to-object maps $O_{src}$ and $O_{ref}$ and $\gamma = 0.2$ is a weighting factor.
Because our rescoring method does not introduce any learnable parameters, we do not need to train our method with the point cloud registration method jointly. 
Therefore, our method can be easily adapted to most feature-based registration methods, bot point-level matching~\cite{fu2021robust} and super-point matching~\cite{yu2021cofinet, predator,qin2022geometric}. 

\subsection{Loss Functions}
We utilize the Negative Cross-Entropy (NCE) loss in its sparse form to supervise the soft correspondence prediction of scene graph matching. Having $\left \| \textbf{S} \right \|$ as the number of nonzero elements of $\textbf{S}$, the scene graph matching loss per sample $\mathcal{L}_{s}$ is defined as:
\begin{equation}
\label{eqn:loss_s}
\mathcal{L}_{s} = \frac{1}{\left \| \textbf{S} \right \|} \sum_{(i, j)}^{\left\{ \textbf{S}_(i,j) \neq 0 \right\}} - \tilde{\textbf{S}}_{i, j} \log(\textbf{S}_{i,j}).
\end{equation}
We compute the ground truth graph similarity $k$ with $k = \left \| \textbf{S} \right \| /  \min(\left | M_{ref} \right |, \left | M_{src} \right |)$ and use Mean Square Error (MSE) loss to supervise the learning of $\tilde{k}$ :
\begin{equation}
    \begin{gathered}
        \mathcal{L}_{k} = \left( k - \tilde{k}\right)^{2}
    \end{gathered}
\end{equation}
With the weighting factor $\alpha = 10$ and the batch size $N$, the overall loss per batch is then:
\begin{equation}
    \mathcal{L} = \frac{1}{N} \sum_{i}^{N} (\mathcal{L}_{s} + \alpha \mathcal{L}_{k}).
\end{equation}

\subsection{Revisiting the Downstream Tasks}
\label{sec:task_rethink}

\noindent \textbf{Overlap Checking} is a direct downstream task of 3D scene graph alignment. 
Sarkar et al.~\cite{sarkar2023sgaligner} proposed to compute a scene-level alignment score $\xi$ representing the percentage of aligned nodes against all nodes in the reference graph. It is reported faster and more accurate than first performing point cloud registration on scene fragments and determining overlapping with the matchability score.

However, we find it is an oversimplified solution to only count the number of scene graph alignments, which may fail to distinguish scene fragments with low-overlapping and non-overlapping. Instead, we frame the problem as measuring the graph similarity between scene graphs. Inspired by the two-stages strategy proposed in SimGNN~\cite{bai2019simgnn}, we jointly consider the coarse global graph similarity score of $k$ and the fine-gained node-level similarity $\tilde{\textbf{S}}$ of all alignment pairs and define the scene-level alignment score $\mu$ as: 
\begin{equation}
    \mu = \tilde{k}\cdot\frac{1}{\left \| \tilde{\textbf{S}} \right \|} \sum_{(i, j)}^{\left\{ \tilde{\textbf{S}}_(i,j) \neq 0 \right\}} \tilde{\textbf{S}}_{(i, j)} \quad.
\label{eqn:overlap}
\end{equation}

\noindent \textbf{Point Cloud Registration.}
The 3D scene graph alignment can be used to reduce the search space of the point-wise matching for the point cloud registration. In SGAligner, the source and reference point clouds are divided into matched object pairs using the estimated graph alignment. Then feature-based point cloud registration is used to search point-wise correspondence traverse through all matched object pairs. Finally, the transformation $T$ is estimated using a robust estimator on all point correspondences. 

\begin{figure}[htb]
    \centering
    \includegraphics[width=0.9\linewidth]{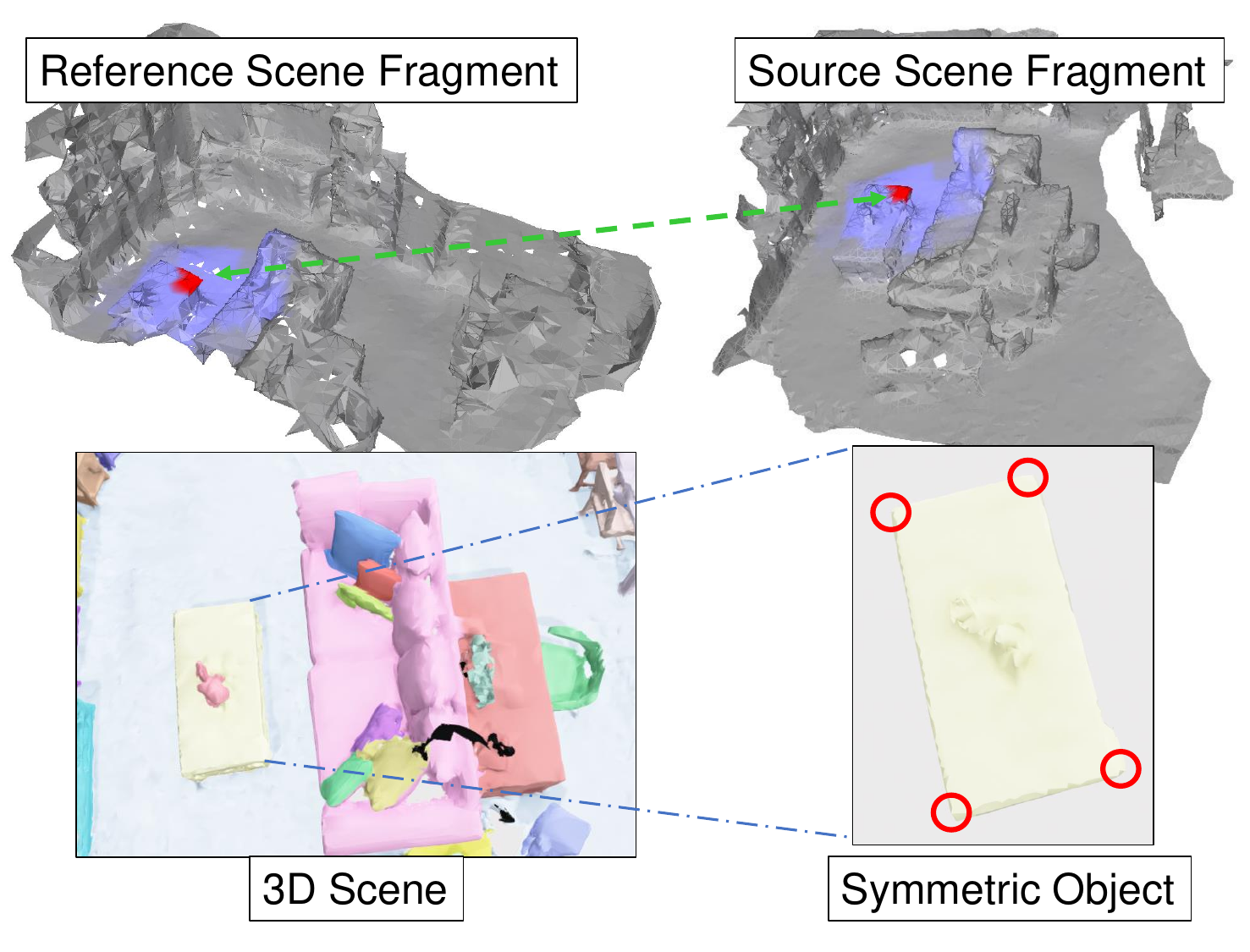}
    \caption{\textbf{Long-range cross-object geometric feature} is gathered in registration method~\cite{qin2022geometric} with transformer. Points in red circles are difficult to match without taking nearby objects as a reference.}
    \label{fig:registraion_transformer}
\end{figure}
Recent point cloud registration methods~\cite{qin2022geometric, yu2023peal} successfully encode long-range geometric context with Transformer~\cite{vaswani2017attention}.  
As visualized in Figure~\ref{fig:registraion_transformer}, matching points (colored red) in symmetric objects or planar objects are under-determined if the reference information from other neighbor objects (colored purple) is missing. Dividing the scene fragment into objects will block access to long-range cross-object geometric features and potentially results in less accurate point matching estimation. 
We simplify this process and use the alignment results to mask out unmatched objects from point clouds and conduct registration on the potential overlapping region only once.

\section{Experiments}
\label{sec:eval}

We evaluate our method for scene graph alignment and overlap-checking (Sec.~\ref{sec:exp_align}), 3D point cloud registration and mosaicking (Sec.~\ref{sec:exp_regis}) and provide an ablation study (Sec.~\ref{sec:eval_ab}). 
For alignment and registration tasks, we follow the data prepossessing method in~\cite{sarkar2023sgaligner} and generate 15,277 training samples and 1,882 validation samples from the 3RScan dataset~\cite{wald2019rio, wald2020learning}. 
Sample numbers are different from the original data splits, due to \textit{the uncontrolled random seed} in their implementation. 
For evaluating the overlap-checking, another 1,882 non-overlap sample pairs are added to the validation subset. 

In the following experiments, we ran SGAligner on our generated data splits and marked the results as \textbf{SGA*} and listed the results of SGAligner in the original paper as reference. 
We pick GeoTransformer~\cite{qin2022geometric} pretrained on 3DMatch~\cite{zeng20163dmatch} for registration and use its KPConv~\cite{thomas2019kpconv} backbone to extract geometric embedding.
For ablation study, we incrementally add our proposed modules to our baseline \textbf{B} graph matching network: (1) \textbf{B+P} as adding P2SG Fusion, (2) \textbf{B+P+K} as adding Soft-topK and AFA-U, (3) \textbf{SG-PGM (B+P+K+S)} as adding Super-point Matching Rescoring, (4) \textbf{SG-PGM+R} as using Graph-Cut RANSAC~\cite{barath2018graph} for pose estimation. 
Implementation details and evaluation metrics definitions are in Appendix~\ref{sec:imp_detail} and~\ref{sec:metrics}.

\subsection{Scene Graph Alignment and Overlap Checking}
\label{sec:exp_align}
We initially evaluate our method for aligning 3D scene graphs using metrics from~\cite{sarkar2023sgaligner}. However, metrics like Hits@k and Mean Reciprocal Rank (MeanRR) do not account for false-positive matches. Therefore, we also assess our results using the F1-score (the harmonic mean of the precision and recall).
We ignore Intra-Graph Alignment Recall metric (IGAR) because "self-aligned" is by design not allowed in our method.

\begin{table}[htb]
\centering
\scalebox{0.9}{
\begin{tabular}{l| c |c |c c c }
\hline
\multirow{2}{*}{\textbf{Methods}} & \multirow{2}{*}{\shortstack{\textbf{Mean}\\ \textbf{RR}} } &  \multirow{2}{*}{\textbf{F1}} & \multicolumn{3}{c}{\textbf{Hits} @ } \\
 & & &  \textbf{K=1} & \textbf{K=3} & \textbf{K=5} \\ \hline  \hline
 
SGA~\cite{sarkar2023sgaligner} & 95.0 & - & 92.3 & 97.4 & 98.7 \\ \hline
SGA* & 96.3 & \underline{89.3} & 94.3 & 96.9 & 98.0 \\ \hline \hline
B   & 89.6 & 62.9 & 82.2 & 98.4 & 99.2  \\ \hline
B+P' (PointNet)   & 97.5 & 79.1 & 95.6 & 99.5 & 99.8  \\ \hline
B+P (KP-Conv)   & \textbf{98.7} & 79.0 & \textbf{97.7} & \underline{98.7} & \underline{99.9}  \\ \hline
B+P+K w/o AIS & 94.2 & 81.8 & 90.1 & 97.0 & 98.4  \\ \hline
B+P+K w/ AIS  & \underline{98.6} & \textbf{89.4} & \underline{97.5} & \textbf{99.7} & \textbf{99.9}  \\ \hline
\end{tabular}
}
\caption{\textbf{Evaluation on node matching.} We evaluate the scene graph node alignment of our method's different variants and compare it with SGAligner. All metrics are the-higher-the-better.}
\label{tab:sgm}\vspace{-2mm}\end{table}

As shown in Table~\ref{tab:sgm}, adding the proposed P2SG Fusion to the baseline significantly improves the node alignment accuracy and is already higher than SGAligner.
With the Soft-topK module, our method can also effectively surpass the false-positive matching pairs and therefore yield the highest F1 score. 
This is not only important for scene graph alignment but also reduces the inference operations (from an average of 16.6 pairs per sample in the validation set to 12.8 pairs) if we later want to conduct the registration in an object-per-object fashion. Furthermore, we modified our network by using PointNet~\cite{qi2017pointnet} as the geometric feature extractor (B+P') for fair comparison with SGAligner, and justify the effectiveness of the AIS Module against simple matrix product for computing node-wise feature similarity (w/o AIS). Since Superpoint Matching Rescoring is only used during registration and has no impact on alignment, we ignore that variant here. 
Results of alignment on the predicted 3D scene graph are given in Appendix Table~\ref{tab:more_sgm_results}.

\begin{figure}[htb]
    \centering
    \includegraphics[width=\linewidth]{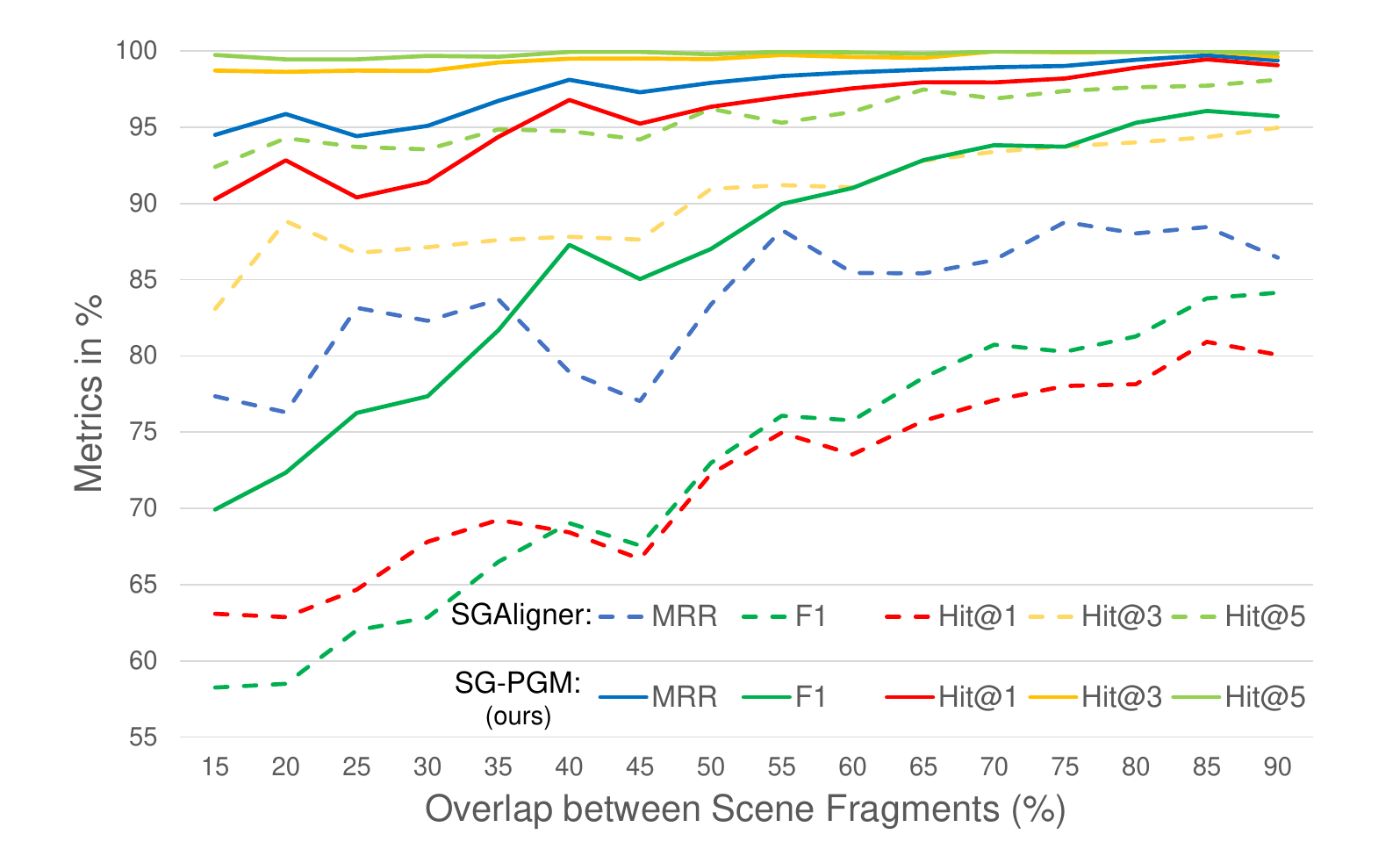}
    \caption{\textbf{Evaluation on node matching with transformation} $T\neq I_{4}$. Results are distributed per overlap range.}
    \label{fig:sgm_distri}
    \vspace{-2mm}
\end{figure}

We provide a more practical evaluation by augmenting random transformation between two scene fragments, different from the $T=I_{4}$ benchmark in~\cite{sarkar2023sgaligner}. 
We trained SGAligner with random $T$ and Gaussian noise as augmentation (SGA*). 
The results are divided into different scene overlapping ranges in Figure~\ref{fig:sgm_distri}. Even though retrained with augmentation, SGAligner still shows a significant accuracy drop compared to results in Table~\ref{tab:sgm}, while the overall performance of our method drops only slightly. 
This demonstrates that fusing graphs and geometric features with our method is robust against rotation. 
However, the F1 score of our method in the overlapping range 10-30\% is more than 20\% lower than in the high-overlap case (see Table~\ref{tab:sgm_distri} in the Appendix).
This means that our method provides more false-positive matching in
low-overlap cases compared to high-overlap cases, but still much better than~\cite{sarkar2023sgaligner}. 

\noindent \textbf{Overlap check of two scene fragments.} To check overlaps, we report scene fragment pairs with the scene-level alignment score $\mu < 0.375 $ as non-overlapped scenes in Table~\ref{tab:sgm_overlap}. As mentioned, our method provides more false-positive results in low-overlapping scenarios. Therefore, we suggest using the top 3 of $\tilde{\textbf{S}}$ scores instead of all $\tilde{\textbf{S}}$ in Eq.~\ref{eqn:overlap} and report non-overlapping with ${\mu_{3}}' < 0.45 $. This variant (SG-PGM@3) suppresses the impact of false-positive node alignments and yields better performance. 
We also analyze the confusion between low-overlap and non-overlap. 
While SGAligner predicts about \textbf{36\%} of low-overlap samples as non-overlap, our method (Ours@3) predicts only \textbf{16\%} of low-overlap samples incorrectly. An extended experiment on overlap-checking is given in Appendix Table~\ref{tab:sgm_overlap_more}.

\begin{table}[htb]
\centering
\scalebox{0.9}{
    \begin{tabular}{l|c c c} \hline
       \textbf{Methods}   & \textbf{Prec.} & \textbf{Recall} & \textbf{F1}  \\ \hline \hline
       SGA~\cite{sarkar2023sgaligner} & 92.03 &  90.94 & 91.48\\ \hline
       SGA* & 93.29 &  90.34 & 91.79\\ \hline
       SG-PGM \small{(ours)} & \underline{94.59} &  \underline{92.03} & \underline{93.29}\\ \hline
      SG-PGM@3 \small{(ours)} & \textbf{95.41} & \textbf{95.01} & \textbf{95.21} \\ \hline
    \end{tabular}
    }
    \caption{\textbf{Overlap check for point cloud registration.} $T=I_{4}$ between fragments. All metrics are the-higher-the-better.}
    \label{tab:sgm_overlap}\vspace{-2mm}
\end{table}

\subsection{Point Cloud Registration and Mosaicking}
\label{sec:exp_regis}
In this section, we use the scene graph alignment result from SGAligner and our method's variants as priors, to support pretrained GeoTransformer~\cite{qin2022geometric} for point cloud registration and mosaicking. 
We evaluate the registration accuracy with Chamfer Distance (CD), Relative Rotation and Translation Error (RRE and RTE), Feature Matching Recall (FMR) and Registration Recall (RR). 
As shown in Table~\ref{tab:regis}, our method outperforms SGAligner in 4 out of 5 metrics even without a robust estimator (Ours+R). Our registration strategy is also 4 times faster than SGAligner. 
\begin{table}[htb]
\centering
\scalebox{0.875}{
    \centering
    \begin{tabular}{l|c c c | c c} \hline
      \textbf{Methods} & \textbf{CD} & \textbf{RRE} & \textbf{RTE} & \textbf{FMR} & \textbf{RR} \\ \hline \hline
       GeoTr~\cite{qin2022geometric} & 0.0312 & 2.3726 & 4.14 & 98.50 & 98.37 \\ \hline
       SGA~\cite{sarkar2023sgaligner} & 0.0111 & 1.012 & 1.67 & \textbf{99.85} & 99.40 \\ \hline
       SGA* & 0.0130 & 1.2929 & 2.11 & \underline{99.74} & 98.87 \\ \hline
       SG-PGM \small{(ours)} & \textbf{0.0083} & \underline{0.6252} & \underline{1.32} & 99.73 & \textbf{99.57} \\ \hline
       SG-PGM+R \small{(ours)}  & \underline{0.0102} & \textbf{0.5103} & \textbf{1.27} & 99.73 & \underline{99.47} \\ \hline
    \end{tabular}
    }
    \caption{\textbf{3D point cloud registration} with $T=I_{4}$. Graph-Cut RANSAC~\cite{barath2018graph} is used in "Ours+R" and SGAligner. FMR and RR: higher-the-better. Others: lower-the-better.} 
    \label{tab:regis}\vspace{-2mm}
\end{table}

We further increase the difficulty of registration and augment the point clouds with random transformation $T$, as shown in Table~\ref{tab:regis_ab}. 
The aim of scene graph alignment before registration is to filter non-overlap parts and encourage point matching within object pairs.
We design the Semantic Consistency of Point Correspondence (SCC) metric, to measure the consistency between predicted point pairs and the ground truth scene graph node pairs:
\begin{equation}
\label{eqn:corr_s}
    \begin{gathered} 
    SCC = \frac{1}{\left | \textbf{C} \right |} \sum_{(i,j)}^{(i,j)\in \textbf{C}} f(i, j) \quad, \\
     f(i, j) =
     \begin{cases} 
       1 & \text{if }  O_{src}(j) = \textbf{S}(O_{ref}(i)) \\
       0       & \text{if } O_{src}(j) \neq \textbf{S}(O_{ref}(i))
        \end{cases},
    \end{gathered}
\end{equation}
in which $\textbf{C}$ is the point-level matching between the reference and source point cloud. $O$ is the point-to-object map and $\textbf{S}$ is the ground truth scene graph alignment.

\begin{table}[htb]
\centering
\scalebox{0.875}{
    \begin{tabular}{c|c | c c | c c c } \hline
      \textbf{Mtds.} & \textbf{Overlap}& \textbf{RRE} & \textbf{RTE} & \textbf{FMR} & \textbf{RR} & \textbf{SCC} \\ \hline  \hline
       \parbox[t]{2mm}{\multirow{4}{*}{\rotatebox[origin=c]{90}{GeoTr~\cite{qin2022geometric}}}}   
       & 10-30 & 8.2130 & 19.40 & 92.47 & 92.73 & 76.98\\
       & 30-60 & 0.4584 & 1.53 & 99.76 & 99.76 & 88.68\\
       & 60 - & 0.2126 & 1.02 & 100.0 & 99.85 & 90.34\\  \cline{2-7}
       & overall  & 1.9398 & 4.96 & 98.37 & 98.37 & 86.90 \\ \hline
        \parbox[t]{2mm}{\multirow{4}{*}{\rotatebox[origin=c]{90}{B+P}}}   
       & 10-30 & 10.169 & 22.53 & 93.33 & 91.11 & 78.98 \\
       & 30-60 & 0.6513 & 1.56 & 99.75 & 99.63 & 90.51\\
       & 60- & 0.1594 & 0.65 & 100.0 & 99.86 & 91.24 \\  \cline{2-7}
       & overall  &  2.2864 & 5.23  &  98.62 & 98.09 & 88.58  \\ \hline
        \parbox[t]{2mm}{\multirow{4}{*}{\rotatebox[origin=c]{90}{B+P+K}}}   
       & 10-30  & 8.9309 & 19.04 & 94.44 & 92.22 &  81.85\\
       & 30-60 & 0.2597 & 0.90  & 99.75 & 99.75 &  91.15\\
       & 60-   & 0.1598 & 0.66 & 100.0 & 100.0 &  91.67 \\ \cline{2-7}
       & overall    & \underline{1.8807} & \underline{4.28}  & \underline{98.83} & \underline{98.41} &  \underline{89.57}\\\hline
        \parbox[t]{2mm}{\multirow{4}{*}{\rotatebox[origin=c]{90}{\shortstack{SG-PGM\\ \small{(ours)}} }}}   
       & 10-30  & 7.3368 & 15.24 & 97.22 & 93.61 & 87.60\\
       & 30-60 & 0.2419 & 0.86  & 100.0 & 99.88 & 93.66\\
       & 60-   & 0.1564 & 0.60 & 100.0 & 100.0 & 93.89 \\ \cline{2-7}
       & overall    & \textbf{1.5668} & \textbf{3.51}  & \textbf{99.47} & \textbf{98.72} & \textbf{92.59} \\\hline
    \end{tabular}
     }
    \caption{\textbf{3D point cloud registration per overlap.} Random transformation is augmented to the scene fragments.  Comparison against GCNet~\cite{zhu2022leveraging} is in Appendix Table~\ref{tab:regis_more}.
    }
    \label{tab:regis_ab}\vspace{-4mm}
\end{table}

More accurate scene graph alignment can filter more non-overlapped objects and reduce the search space of the registration method. 
It explains the accuracy improvement from the B+P variant to the B+P+K variant of our method. 
Without our Superpoint Rescoring Method, the registration accuracy in low-overlap cases (10-30\%) is merely better than~\cite{qin2022geometric}, though the overall performance is better. 
After adding the Superpoint Rescoring Method, our complete pipeline shows the best performance in all overlapping ranges, especially improving SCC with a large margin. 
This shows the effectiveness of guiding point matching with semantic priors in low-overlap scenarios. 

\noindent \textbf{Point cloud mosaicking} is the task of registering a set of partial point clouds to reconstruct the completed scene. 
As proposed in~\cite{sarkar2023sgaligner}, the mosaicking is conducted by running pairwise registration for all pairs. 
We select 143 scenes for testing point cloud mosaicking and the results are listed in Table~\ref{tab:regis_mosaick}. 
We use the same metrics as in~\cite{sarkar2023sgaligner} to evaluate the results: accuracy and completeness of the resulting reconstruction (the-lower-the-better), precision, recall, and F1-score of registered point clouds (the-higher-the-better). 
As expected, our method shows higher accuracy than others. Qualitative results of registration and mosaicking are given in Appendix~\ref{sec:vis}.

\begin{table}[htb]
\centering
\scalebox{0.85}{
    \begin{tabular}{l|c c | c c c} \hline
      \textbf{Methods} & \textbf{Acc} & \textbf{Comp} & \textbf{Prec} & \textbf{Recall} & \textbf{F1} \\ \hline  \hline
       GeoTr~\cite{qin2022geometric}  & 0.1213 & 0.0917 & 95.84 & 87.17 & 90.11 \\ \hline
       SGA~\cite{sarkar2023sgaligner} & 0.0094 & 0.0935 & 90.87 & 97.44 & 93.58 \\ \hline
       SG-PGM \small{(ours)} & \underline{0.0033} & \underline{0.0040} & \underline{99.81} & \underline{99.79} & \underline{99.80}\\ \hline
       SGPGM+R \small {(ours)}  & \textbf{0.0024} & \textbf{0.0026} & \textbf{99.86} & \textbf{99.85} & \textbf{99.86} \\ \hline
    \end{tabular}
    }
    \caption{\textbf{Point cloud mosaicking from multiple fragments.} Our method outperforms others even without using RANSAC.}
    \label{tab:regis_mosaick} 
\end{table}

\begin{figure*}[htb]
    \centering
    \begin{subfigure}[b]{0.45\linewidth}
        \includegraphics[width=\textwidth]{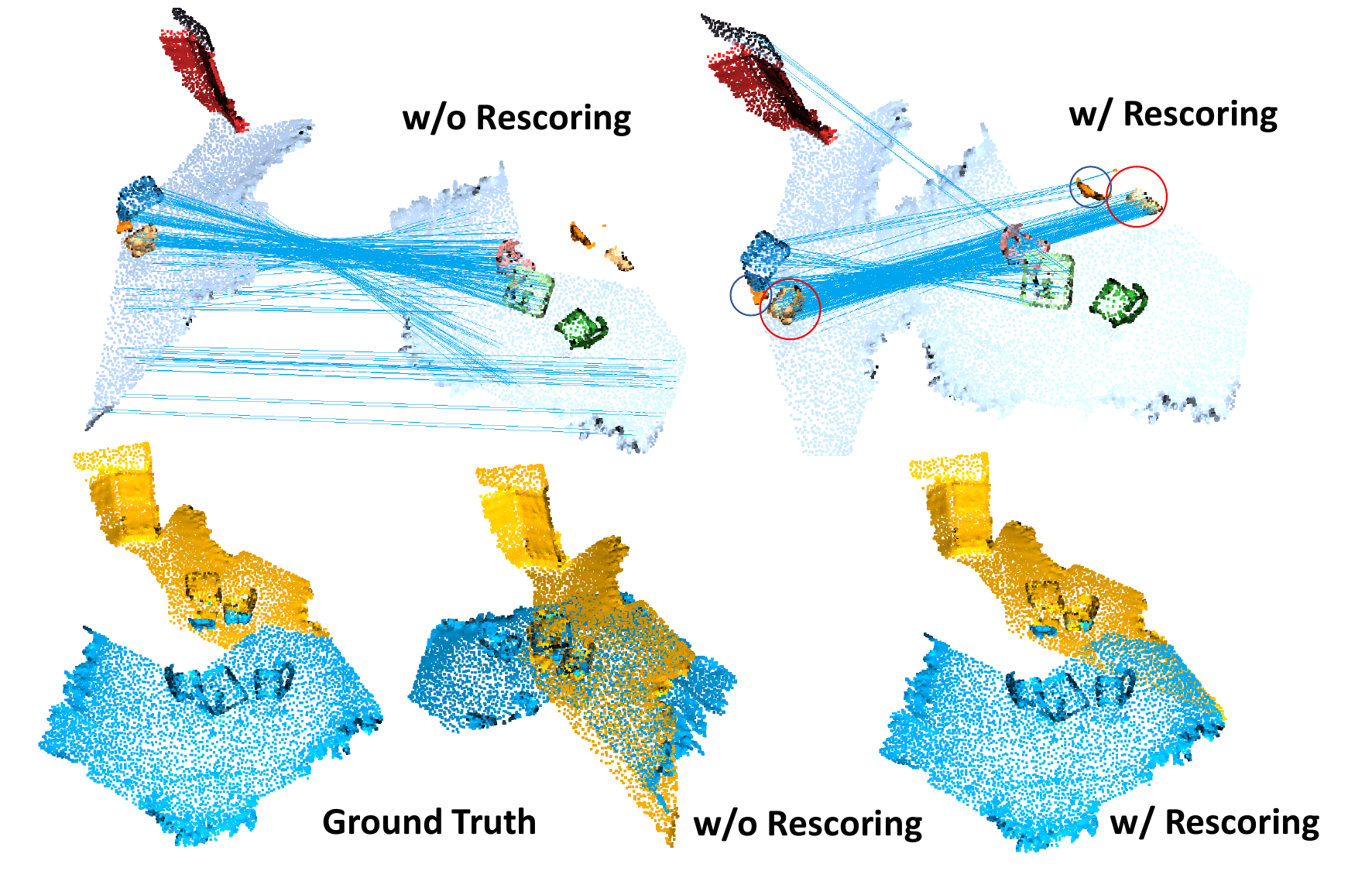}
    \end{subfigure}
    \begin{subfigure}[b]{0.52\linewidth}
        \includegraphics[width=\textwidth]{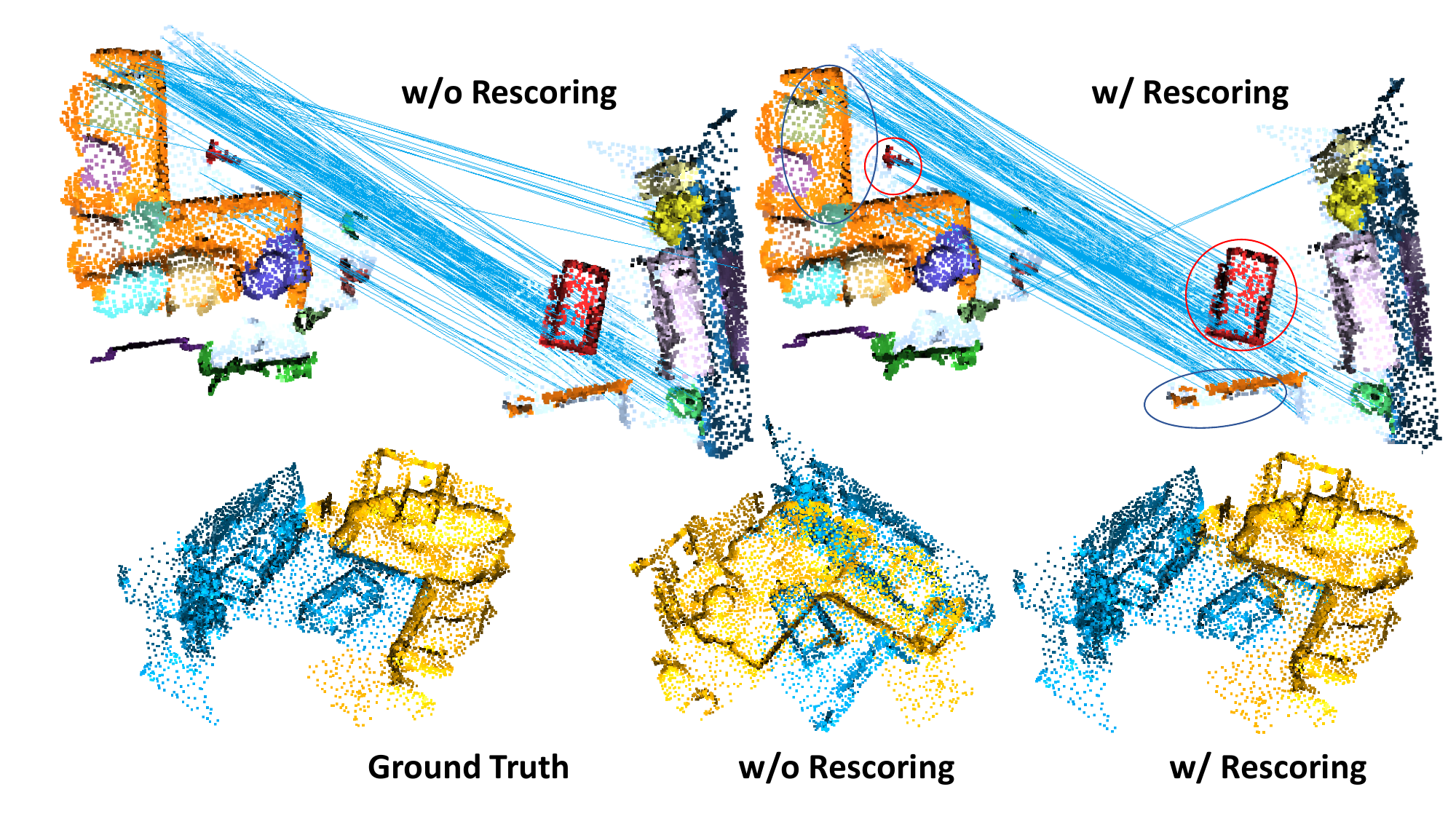}
    \end{subfigure}
    \caption{\textbf{Registration results with and without Superpoint Matching Rescoring of low overlapping scene fragments.}}
    \label{fig:rescore}\vspace{-2mm}
\end{figure*}

\subsection{Aligning 3D Scenes with Changes}
\label{sec:eval_changes}
3RScan dataset provides multiple rescans of one scene with changes such as moved, removed, and deformed objects. 
Following SGAligner~\cite{sarkar2023sgaligner}, we investigate the alignment in the following scenarios: (i) aligning a sub-scene on the original scan that contains no changes; (ii) aligning a 3D sub-scene on a rescan that contains changes; and (iii) aligning sub-scenes that contains changes.

\begin{table}[htb]
\centering
\scalebox{0.9}{
\begin{tabular}{c| c |c |c c c }
\hline
\multirow{2}{*}{\textbf{Methods}} & \multirow{2}{*}{\textbf{Dynamics.}} & \multirow{2}{*}{\textbf{MRR}}  & \multicolumn{3}{c}{\textbf{Hits} @ } \\
  & & &  \textbf{K=1} & \textbf{K=3} & \textbf{K=5} \\ \hline  \hline
\multirow{3}*{\shortstack{SGA*\\ \textbf{\cite{sarkar2023sgaligner}}}} 
& (i)    & 97.9  & 96.6 & 99.1 & 99.7 \\ 
& (ii)   & 93.6  & \textbf{90.6} & 96.2 & 97.5 \\
& (iii)  & 88.8  &87.1 & 94.2 & 96.2 \\\hline
\multirow{3}*{\shortstack{SG-PGM\\ \small{(ours)}}} 
& (i)    & \textbf{99.8} & \textbf{99.7} & \textbf{99.9} & \textbf{100} \\ 
& (ii)  & \textbf{94.2} &   90.0 & \textbf{98.2} & \textbf{99.3} \\
& (iii) & \textbf{93.4} &   \textbf{88.9} & \textbf{97.7} & \textbf{99.2} \\\hline
\end{tabular}
}
\caption{\textbf{Alignment of a local 3D scene to a prior 3D map with differences in overlap and changes.}}
\label{tab:sgm_change}\vspace{-2mm}\end{table}

We run SGAligner on our generated data samples and list the results together with ours in Table~\ref{tab:sgm_change}. Our approach outperforms SGAligner in most metrics of all three scenarios, which indicates the strong robustness to scene changes. Details about the data generation of these scenarios and extra experiments against various controlled semantic noises are in Appendix~\ref{sec:imp_detail} and~\ref{sec:eval_noise}.

\subsection{Ablation Study}
\label{sec:eval_ab}
We focus only on the rescoring method and the registration strategy. 
Since the effectiveness of the partial graph matching (SOFT-topK) and feature fusion (P2SG) module has been evaluated and verified in Section~\ref{sec:exp_align} and~\ref{sec:exp_regis}.

\noindent \textbf{Super-point Matching Rescoring}
As shown in Table~\ref{tab:regis_ab}, with the help of the Super-point Matching Rescoring, our method shows obviously better performance in terms of \textit{SCC} compared to other variants. We visualize the point cloud registration results of our method with or without using the Super-point Matching Rescoring method in Figure~\ref{fig:rescore}. From that, we obverse that rescoring the point matching with scene graph alignment as prior, can avoid mismatching of points with similar local geometric features but belonging to a different object or semantic class.

\noindent \textbf{Registration Strategy}
We build up an experiment to evaluate the registration performance on the same validation split used in~\ref{sec:exp_regis} using the ground truth scene graph alignment and run registration with all-to-all (A2A), object-per-object (OPO) and overlap-to-overlap (O2O) fashions. Results in Table~\ref{tab:regis_dummy} indicate that masking the scene fragments with the perfect overlap region (GeoTr+O2O) yields the best result while traversing through object pairs performs the worst. This also supports our analysis in~\ref{sec:task_rethink}. 

\begin{table}[htb]
\centering
\scalebox{0.9}{
    \begin{tabular}{l|c c c c c} \hline
      \textbf{Methods} & \textbf{RRE} & \textbf{RTE} & \textbf{FMR} & \textbf{RR} \\ \hline \hline
       GeoTr~\cite{qin2022geometric}+A2A   & 1.9398 & 4.96  & 98.37  & 98.37 \\
       GeoTr+OPO   & \underline{5.9528} & \underline{15.46} & \underline{99.63}  & \underline{94.69} \\
       GeoTr+O2O   & \textbf{1.4443} & \textbf{3.67}  & \textbf{99.32}  & \textbf{99.05} \\ \hline
    \end{tabular}
    }
    \caption{\textbf{Ablation study on different registration strategies.}}
    \label{tab:regis_dummy}\vspace{-3mm}
\end{table}

\section{Conclusion}
\label{sec:conlusion}

We have presented SG-PGM, a graph neural network for scene graph partial matching. 
We revisited the geometric feature extraction, partial matching mechanism, and strategies for solving downstream tasks of the existing work~\cite{sarkar2023sgaligner}. 
We designed our method to use more expressive geometric features with the point to scene graph fusion module. 
We proposed the Super-point Rescoring method for boosting point cloud registration with semantic priors. 
Compared to the existing work~\cite{sarkar2023sgaligner, qin2022geometric}, our method shows significant performance improvements on scene graph alignment, overlap-checking, point cloud registration, and other downstream tasks. 
Moreover, our scene graph alignment method remains decoupled from registration and robust to scene dynamics and noises. 
For future work, we would like to explore the approach for using semantic priors from scene graph alignment to design efficient sparse transformers for geometric feature analysis.

\noindent \small \textbf{Acknowledgement}: The research leading to these results has been partially funded by the German Ministry of Education and Research (BMBF) under Grant Agreement 01IW20009 (RACKET) and the EU Horizon Europe Framework Program under Grant Agreement 101058236 (HumanTech).
\clearpage

{
    \small
    \bibliographystyle{ieeenat_fullname}
    \bibliography{main}
}

% WARNING: do not forget to delete the supplementary pages from your submission 
\clearpage
\appendix
\setcounter{page}{1}
\maketitlesupplementary
\setcounter{section}{0}
\setcounter{equation}{0}
\renewcommand{\thesection}{\Alph{section}}

\begin{abstract}
In the supplemental material, we provide additional details about the following:
\begin{itemize}
    \item Details on implementation. (Section~\ref{sec:imp_detail}),
    \item Evaluation metrics of 3D scene graph alignment and downstream tasks (Secion~\ref{sec:metrics}),
    \item Evaluation on scene graph alignment with controlled semantic noise and with predicted 3D scene graph (Section~\ref{sec:eval_noise}),
    \item Additional ablation study on registration strategy and network variants (Section~\ref{sec:more_ablation}),
    \item Visualisation on point cloud registration and point cloud mosaicking (Section~\ref{sec:vis}).
\end{itemize}

\end{abstract}

\section{Implementation Details}
\label{sec:imp_detail}

\noindent\textbf{Data Generation for Alignment in Dynamics:} 
To evaluate scene graph alignment in the changing environment Section~\ref{sec:eval_changes}, we generate the samples using the sub-scenes in the validation split and the original 3D scene maps from~\cite{wald2020learning, wu2021scenegraphfusion}. 
The dynamics between scan and rescan of the same indoor scene consist of three types: "non-rigid", "removed" and "rigid". 
We ignore small rigid object changes, whose Euler angles $\alpha + \beta + \gamma < 3^{\circ}$, and mark them as aligned node ground truth. 
Thus, the sample numbers of scenarios (i), (ii) and (iii) are 819, 354, and 1,635.

\noindent\textbf{Network and Training:} We take the fine-level geometric feature of the KPConv-FPN as the input of our P2SG Fusion module.
Same as suggested in~\cite{wang2023deep}, the input node embeddings of the AFA-U module are set to \textbf{zero vectors} for one graph and \textbf{one-hot vectors} for the other graph. 
Unlike in~\cite{wang2023deep}, we train the AFA-U module together with the other parts of the network in one stage.
We employ the matching rescoring on the super-point matching stage of~\cite{qin2022geometric} because the fine-level points within a super-point are considered most likely to belong to the same object. 
The training procedure takes 10 epochs with the ADAM optimizer and an initial learning rate of $1e^{-4}$, which decreases by 0.1 every 4 epochs. 
If not specified, we mask out the unmatched objects of the scene fragments and conduct registration on the overlap region as a whole instead of registration traverse through all matched pairs.

\section{Evaluation Metrics}
\label{sec:metrics}
We give the definition of evaluation metrics used in the main paper here. For the same evaluation metric used in multiple tasks, its definition will be adjusted based on input.

\subsection{Scene Graph Alignment}

\textbf{Hits@K} describes the fraction of true entities that appear in the first $k$ entities of the sorted rank list $R$ of the alignment prediction $\tilde{\textbf{S}}$. Denoting the set of individual ranks as $r_{i}$, it is given as:
\begin{equation}
    H_{k}(r_1, ..., r_n) = \frac{1}{n} \sum^{n}_{i} \left[ r_i < k \right]  \quad \in \left[0, 1 \right]
\end{equation}
where $[\cdot]$ is the Iversion bracket.

\textbf{Mean Reciprocal Rank (MRR)} is the arithmetic mean over the reciprocals of ranks of true triples:
\begin{equation}
    MRR(r_1, ..., r_n) = \frac{1}{n} \sum^{n}_{i} \frac{1}{r_i} \quad \in \left(0, 1 \right]
\end{equation}

\textbf{F1-score} is the harmonic mean of the precision and recall. More specifically, the F1 score for graph matching is defined as:
\begin{equation}
    \begin{gathered}
        tp,\: fp,\: fn =  \tilde{\textbf{S}} \textbf{S}, \: \tilde{\textbf{S}} (1- \textbf{S}), \:(1-\tilde{\textbf{S}} ) \textbf{S}\\
        F1 = \frac{2tp}{2tp+fp+fn}  \quad \in \left[0, 1 \right].
        \end{gathered}
\end{equation}

\subsection{Overlap Checking}
Overlap checking of two 3D scenes is a binary classification problem that checks whether two 3D scenes overlap or not. Metrics (Precision, Recall, and F1-score) are given as:
\begin{equation}
    \begin{gathered}
        Prec. = \frac{TP}{TP + FP }\quad \in \left[0, 1 \right], \\
        Recall = \frac{TP}{TP + FN}\quad \in \left[0, 1 \right], \\
        F1 = 2\frac{Prec. \times Recall}{Prec. + Recall}\quad \in \left[0, 1 \right],
    \end{gathered}
\end{equation}
in which $TP$ is true positive, $FP$ is false positive and $FN$ as false negative.
\subsection{Point Cloud Registration}
\textbf{Registration Recall (RR)} is the fraction of successfully registered point cloud pairs. A point cloud
pair is successfully registered when its transformation error is lower than threshold $\tau_{1} = 0.2m$.
In addition, the transformation error is the root mean square error of the ground truth correspondence $C$, to which the estimated transformation $\tilde{\textbf{T}}$ has applied:
\begin{equation}
    \begin{gathered}
        RMSE = \sqrt{\frac{1}{\left | C \right |} \sum_{(p_{x}, q_{y})\in C} \left \|  \tilde{\textbf{T}} (\textbf{p}_x) - \textbf{q}_{y} \right \|^{2}_{2} } ,\\
        RR = \frac{1}{M} \sum_{i=1}^{M} \left[RMSE < \tau_{1} \right] \quad \in \left[0, 1 \right],
    \end{gathered}
\end{equation}
where $p_x$ and $q_y$ denote the $x$-th point in source $P$ and $y$-th point in reference $Q$, respectively; $[\cdot]$ is the inerson bracket; and $M$ is the number of all point cloud pairs.

\textbf{Feature Matching Recall (FMR)} is the fraction of point cloud pairs whose Inlier Ration (IR) is above $\tau_{3} = 0.05$. FMR measures the potential success during the registration, while Inlier Ratio is the fraction of inlier correspondences among all hypothesized correspondences $\tilde{C}$:
\begin{equation}
    \begin{gathered}
        IR = \frac{1}{\left | \tilde{C} \right |} \sum_{(p_{x}, q_{y})\in \tilde{C}}  \left[\left \|  \textbf{T} (\textbf{p}_x) - \textbf{q}_{y} \right \|_{2} < \tau_{2} \right] \quad \in \left[0, 1 \right],\\
        FMR = \frac{1}{M} \sum_{i=1}^{M} \left[IR > \tau_{3} \right] \quad \in \left[0, 1 \right], 
    \end{gathered}
\end{equation}
in which an inlier is defined as the distance between the two points is lower than a certain threshold $\tau_{2}$ under the ground-truth transformation $\textbf{T}$.

\textbf{Relative Rotation Error (RRE)} measures the geodesic distance in degrees between the estimated $\tilde{\textbf{R}}$ and ground truth rotation $\textbf{R}$ matrices:
\begin{equation}
    RRE = \arccos(\frac{trace(\textbf{R}^{T}\tilde{\textbf{R}}) - 1)}{2}).
\end{equation}

\textbf{Relative Translation Error (RTE)} measures the Euclidean distance between the estimated $\tilde{\textbf{t}}$ and ground truth translation $\textbf{t}$ vectors:
\begin{equation}
    RTE = \left \|  \textbf{t} - \tilde{\textbf{t}} \right \|.
\end{equation}

\textbf{Modified Chamfer Distance} measures the
average of the pair-wise nearest distance between two point sets $P$ and $Q$: 
\begin{equation}
\begin{gathered}
    CD = \frac{1}{\left| P \right|}\sum_{p\in P} \min_{q\in Q}\left \|  \tilde{\textbf{T}} (\textbf{p}) - \textbf{q}\right \|^{2}_{2} + \\ \quad \quad \quad \quad \frac{1}{\left| Q \right|}\sum_{q\in Q} \min_{p \in P} \left \|  \textbf{q} - \tilde{\textbf{T}} (\textbf{p})\right \|^{2}_{2}
\end{gathered}
\end{equation}

\subsection{Point Cloud Mosaicking}
Having the ground truth point cloud $P$ and reconstructed point cloud $P^{*}$. The \textbf{Reconstruction Accuracy (Acc)} and \textbf{Reconstruction Completeness (Comp)} are defined as:
\begin{equation}
    \begin{gathered}
        Acc = \frac{1}{n}\sum_{p\in P}^{n} \min_{p^{*} \in P^{*}} (\left\| p - p^{*} \right \|) \\
        Comp = \frac{1}{n}\sum_{p^{*} \in P^{*}}^{n} \min_{p\in P} (\left\| p - p^{*} \right \|) 
    \end{gathered}
\end{equation}

And the \textbf{Reconstruction Precision (Prec.)} and \textbf{recall (Recall)} and the \textbf{F1-score} are defined as:
\begin{equation}
    \begin{gathered}
        Prec. =  \frac{1}{n}\sum_{p\in P}^{n} \min_{p^{*} \in P^{*}} \left[\left\| p - p^{*} \right \| < 0.05 \right] \quad \in \left[0, 1 \right],\\
        Recall = \frac{1}{n}\sum_{p^{*} \in P^{*}}^{n} \min_{p\in P} \left[\left\| p - p^{*} \right \| < 0.05 \right] \quad \in \left[0, 1 \right],\\
        F1 = 2\frac{Prec. \times Recall}{Prec. + Recall} \quad \in \left[0, 1 \right].
    \end{gathered}
\end{equation}

\section{Evaluation on Scene Graph Alignment with Controlled Semantic Noise and with Predicted 3D Scene Graph}
\label{sec:eval_noise}
We also test the robustness of our network against controlled noise on scene graph node alignment. Following the same implementation of SGAligner~\cite{sarkar2023sgaligner}, we evaluate our method with 5 different types of noises: (i) only relationships are removed; (ii) only object(node) are removed their corresponding attributes and any relationships that include them are also removed; (iii) both relationships and object nodes are removed; (iv) object instances assigned with the wrong semantic label); and (v) both relationships and objects are both assigned with wrong semantics. Results are given in Table~\ref{tab:appendix_changes}. We also list the noise-free result here as a reference. 

\begin{table}[htb]
\centering
\scalebox{0.9}{
\begin{tabular}{c| c |c |c c c }
\hline
\multirow{2}{*}{\shortstack{\textbf{Noise}\\ \textbf{Types}} } & \multirow{2}{*}{\shortstack{\textbf{Mean}\\ \textbf{RR}} } &  \multirow{2}{*}{\textbf{F1}} & \multicolumn{3}{c}{\textbf{Hits} @ } \\
 & & &  \textbf{K=1} & \textbf{K=3} & \textbf{K=5} \\ \hline  \hline
(i) &  96.70 & 77.52 &  94.93 &  98.56 &  98.80 \\ \hline
(ii) &  97.81 & 78.41 &  96.02 &  99.69 &  99.94 \\ \hline
(iii) &  96.86 & 77.15 &  94.43 &  99.35 &  99.89 \\ \hline
(iv) &  85.18 & 69.71 & 77.99  & 90.69  &  94.75 \\ \hline
(v) &  85.14 & 69.05 & 77.81 & 90.57 & 95.02 \\ \hline
noise-free & 97.91 & 88.39 & 96.24 & 99.66 & 99.93 \\ \hline
\end{tabular}
}
\caption{\textbf{Evaluation on node matching with different variants of controlled semantic noise}. }
\label{tab:appendix_changes}\vspace{-2mm}
\end{table}

Our method shows very strong robustness against missing relationships (edges) and missing instances (nodes). In (iv) and (v), wrong instance semantic information shows relatively strong impacts on the alignment performance compared to wrong relationships. 
For testing the use of predicted 3D scene graphs instead of ground truth graphs, we generated predicted 3D scene graphs using~\cite{wald2020learning} and tested our network (only trained on the ground truth) on the alignment task. 
Since the authors of~\cite{sarkar2023sgaligner} did not publish their code or pre-trained model for using predicted 3D scene graph, we \textbf{cannot guarantee a fair comparison} with their results. Table~\ref{tab:more_sgm_results}  reproduces theirs as in [34] compared with \textbf{ours on our validation set}. 

\begin{table}[htb]
\centering
\scalebox{0.9}{
\begin{tabular}{l| c |c |c c c }
\hline
\multirow{2}{*}{\textbf{Methods}} & \multirow{2}{*}{\shortstack{\textbf{Mean}\\ \textbf{RR}} } &  \multirow{2}{*}{\textbf{F1}} & \multicolumn{3}{c}{\textbf{Hits} @ } \\
 & & &  \textbf{K=1} & \textbf{K=3} & \textbf{K=5} \\ \hline  \hline
SGA~\cite{sarkar2023sgaligner}& 88.2 &  - & 83.3 &  91.8 &  95.1 \\ \hline
B+P+K & 95.9 &  86.0 & 93.1 &  98.6 &  99.4 \\ \hline
\end{tabular}}

\caption{\textbf{Evaluation on node matching with predicted graph.}}
\label{tab:more_sgm_results}\vspace{-4mm}
\end{table}

\section{Additional Ablation Study}
\label{sec:more_ablation}

\subsection{Object-per-Object Registration with Ours}
Same as SGAligner~\cite{sarkar2023sgaligner}, we conduct object-per-object point cloud registration following with RANSAC using the scene graph alignment results of our own network. To further improve the robustness of the object-to-object registration, we propose two methods: (1) The dense scene graph alignment result $\textbf{S}$ is first filtered with a confidence threshold $s$, only when the score of object pairs is higher than $s$ will be considered in point cloud registration. If none of the object pairs has a score higher than $s$, all object pairs are taken for registration, and (2) only top-$k$-scored object pairs will be used in registration. We also give the registration results of using our network with overlap-to-overlap (O2O) and using SGAligner (S$^{\star}$.) with O2O as references in Table~\ref{tab:appendix_regis}. Our network combined with OPO registration performs marginally worse than with O2O registration, while for SGAligner the situation is the converse.

\begin{table}[htb]
\centering
\scalebox{0.9}{
    \begin{tabular}{l|c c c | c c} \hline
      \textbf{Methods} & \textbf{CD} & \textbf{RRE} & \textbf{RTE} & \textbf{FMR} & \textbf{RR} \\ \hline \hline
       $s=0$ & 0.0544 & 4.9849 & 12.31 & 99.37 & 96.00 \\ \hline
       $s=0.3$ & 0.0581 & 4.8246 & 12.74 & 99.37 & 95.74 \\ \hline
       $s=0.5$ & 0.0462 & 3.9634 & 9.74 & 99.26 & 96.39 \\ \hline
       $k=3$ & 0.0627 & 5.1250 & 13.61 & 99.37 & 95.95 \\ \hline
       $k=5$ & 0.0514 & 4.7141 & 11.76 & 99.37 & 96.27 \\ \hline
       $k=7$ & 0.0574 & 5.0628 & 12.97 & 99.37 & 95.90 \\ \hline
       O2O  & \textbf{0.0083} & \textbf{0.6252} & \textbf{1.32} &\textbf{99.73} & \textbf{99.57}  \\ \hline
       S$^{\star}$. + O2O & 0.0179 & 1.3428 & 2.67 & 99.26 & 98.95 \\ \hline
    \end{tabular}
    }
    \caption{\textbf{Object-per-Object Point Cloud Registration with our method.} Methods with $s$ represent filter object pairs with confidence scores lower than the threshold, while methods with $k$ take only the top-$k$ object pairs for registration.}
    \label{tab:appendix_regis}\vspace{-3mm}
\end{table}

%We also provide a simplified explanation of how the object-per-object strategy may fail in Figure~\ref{fig:n2n}.

%\begin{figure}[htb]
    %\centering
    %\includegraphics[width=0.9\linewidth]{figs/n2n.pdf}
    %\caption{\textbf{Object-per-object registration on symmetric and planar shapes} using GeoTransformer.}
    %\label{fig:n2n}
%\end{figure}

\subsection{Fusion with Different Levels of Point Feature}
KPConv-FPN~\cite{thomas2019kpconv} provides multi-level point geometric features of a point cloud. In the original implementation of Geotransformer, there are three levels of geometric features: coarse-level $N_{c} \times 1024$, middle-level $N_{m} \times 512$ and fine-level $N_{f} \times 256$. Here we give a comparison of using different levels of geometric features for the P2SG fusion module in terms of 3D scene graph alignment in Table~\ref{tab:appendix_ptfusion}. As the result shows, P2SG fusion with fine-level geometric features performs the best among all listed variants. 

\begin{table}[htb]
\centering
\scalebox{0.9}{
\begin{tabular}{l| c |c |c c c }
\hline
\multirow{2}{*}{\textbf{Methods}} & \multirow{2}{*}{\shortstack{\textbf{Mean}\\ \textbf{RR}} } &  \multirow{2}{*}{\textbf{F1}} & \multicolumn{3}{c}{\textbf{Hits} @ } \\
 & & &  \textbf{K=1} & \textbf{K=3} & \textbf{K=5} \\ \hline  \hline
Coarse &  97.00 & 85.51 &  94.69 &  99.33 & 99.79  \\ \hline
Middle &  97.85 & 87.67  & 96.24 &  99.58 &  99.83 \\ \hline
Fine   & \textbf{98.58} & \textbf{89.39} & \textbf{97.49} & \textbf{99.68} & \textbf{99.90}  \\ \hline
\end{tabular}
}
\caption{\textbf{Evaluation on node matching with different levels of point geometric feature.}}
\label{tab:appendix_ptfusion}\vspace{-2mm}\end{table}

\subsection{Alignment with Augmented Transformation}
Here we provide the 3D scene graph alignment results with augmented $T$ in Table~\ref{tab:sgm_distri} as the complementary of Figure~\ref{fig:sgm_distri}.

\begin{table}[htb]
    \centering
    \scalebox{0.825}{
    \begin{tabular}{c | c| c |c |c c c}
    \hline
        \multirow{2}{*}{\textbf{Mtds.}} & \multirow{2}{*}{\shortstack{\textbf{Overlap} \\ \textbf{$\left ( \% \right )$}} }& \multirow{2}{*}{\shortstack{\textbf{Mean}\\ \textbf{RR}} } &  \multirow{2}{*}{\textbf{F1}} & \multicolumn{3}{c}{\textbf{Hits} @ } \\
         & & & &  \textbf{K=1} & \textbf{K=3} & \textbf{K=5} \\  \hline \hline
        \multirow{4}*{\shortstack{SG-PGM\\(\textit{ours})}} & 10-30 & 94.96 & 74.86 & 91.23 & 98.69 & 99.65\\ 
        & 30-60 & 97.91 & 87.95 & 96.33 & 99.54 & 99.87\\ 
        & 60- & 99.15 & 95.21 & 98.48 & 99.83 & 99.93\\ \cline{2-7}
        & overall  & \textbf{97.81} & \textbf{88.18} & \textbf{96.16} & \textbf{99.49} & \textbf{99.85}\\ \hline
        \multirow{4}*{\shortstack{SGA*\\ \textbf{\cite{sarkar2023sgaligner}}}} 
        & 10-30 & 79.93 & 60.46 & 64.64 & 86.54 & 93.50\\ 
        & 30-60 & 83.20 & 71.84 & 71.25 & 89.61 & 95.28\\  
        & 60-   & 87.24 & 81.05 & 78.01 & 93.75 & 97.48\\ \cline{2-7}
        & overall & 85.92 & 79.46 & 77.69 & 88.07 & 93.71 \\  \hline
    \end{tabular}
    }
    \caption{\textbf{Evaluation of our proposed method on node matching per overlap range.} Even in low-overlap cases, our method still provides accurate alignment results with Hit@1 over 90\%.}
    \label{tab:sgm_distri}\vspace{-2mm}
\end{table}

\subsection{Analyse of AIS Module}
\begin{figure}[htb]
     \centering
     \begin{subfigure}[b]{0.48\linewidth}
         \includegraphics[width=\textwidth]{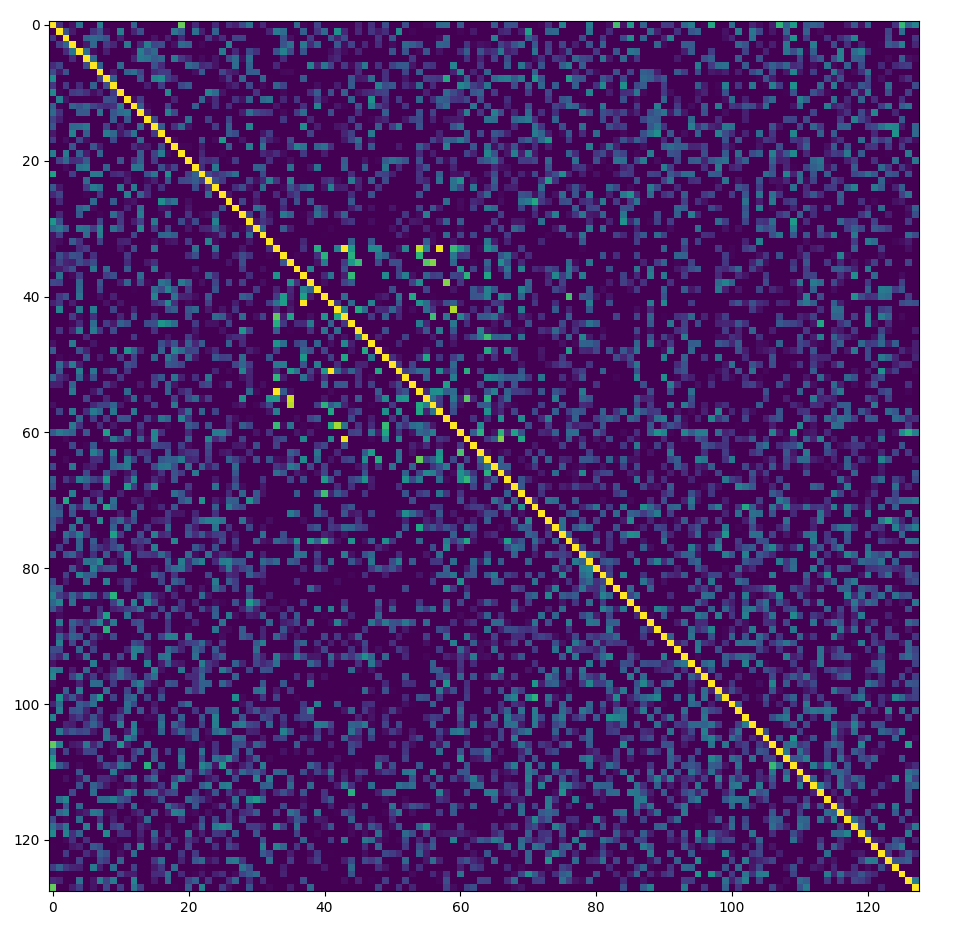}
         \subcaption{$W_{s}$}
     \end{subfigure}
     \begin{subfigure}[b]{0.48\linewidth}
         \includegraphics[width=\textwidth]{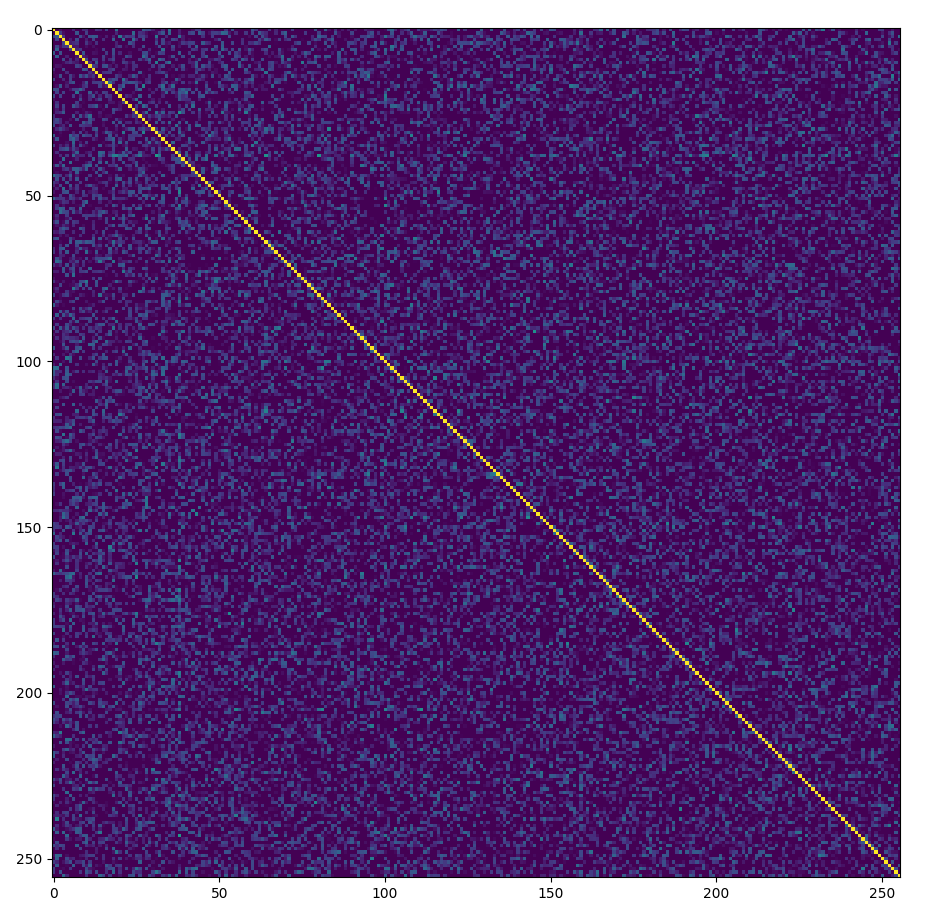}
         \subcaption{$W_{p}$}
     \end{subfigure}
     \caption{\textbf{The learnable parameters $W_{s}$ and $W_{p}$ of the AIS Module.}}
    \label{fig:affi_vis}
\end{figure}

Equation~\ref{eqn:ais} gives the definition of the affinity matrix, in which the affinity of the embeddings from the scene graph and the point cloud is separately computed. 
In Figure~\ref{fig:affi_vis}, we provide a visualization of the learnable parameters $W_{s}$ and $W_{p}$ . 
As shown in the Figure, the multi-level scene graph embedding is more coupled crossing different feature channels, especially of the first-hop graph embedding, while the geometric feature is relatively more decoupled.

\subsection{Additional comparison with GCNet on point cloud registration and overlap checking}
We tested GCNet~\cite{zhu2022leveraging} on the registration task on our validation set in Table~\ref{tab:regis_more}. We additionally combined our method with GCNet to mask out the feature points from unmatched objects before the Consistent Voting, which shows improvement compared to GCNet alone. 

\begin{table}[htb]
\centering
\scalebox{0.9}{
    \centering
    \begin{tabular}{l| c c | c c} \hline
      \textbf{Methods}  & \textbf{RRE} & \textbf{RTE} & \textbf{FMR} & \textbf{RR} \\ \hline \hline
      GeoTr [31] & 1.94 & 4.96 & 98.37 & 98.37 \\ 
       GeoTr + Ours & \textbf{1.57} & \textbf{3.51} & \textbf{99.47} & \textbf{98.72} \\ \hline 
       GCNet \cite{zhu2022leveraging}  & 2.24 & 5.43 & 98.88 & 98.51 \\ 
       GCNet + Ours  & 1.96 & 4.91 & 99.09 & 98.72 \\ \hline 
    \end{tabular}
    }
    \caption{\textbf{Additional evaluation on point cloud registration.}} 
    \label{tab:regis_more}\vspace{-2mm}
\end{table}

We also tested GCNet on the overlap checking task, using the average of the top 25\% of predicted overlap score vector $o$ and saliency score vector $s$. In Table~\ref{tab:sgm_overlap_more}, we report GCNet with $o_{25\%}\cdot s_{25\%} > 0.45$ as overlap, and the results of using the scene-level score $k$ instead of Eq.~\ref{eqn:overlap} in our method. It shows a huge drop in Prec.~because our partial graph matching module is only trained with overlapping samples.
\begin{table}[htb]

\centering
\scalebox{0.9}{
    \begin{tabular}{l|c c c} \hline
       \textbf{Methods}   & \textbf{Prec.} & \textbf{Recall} & \textbf{F1}  \\ \hline \hline
       SGA [34] & 92.03 &  90.94 & 91.48\\ \hline
      GCNet \cite{zhu2022leveraging} & 93.43 &  92.24 & 92.83\\ \hline
      SG-PGM w/ $k>0.45$  & 89.94 &  96.87 & 93.28 \\ \hline
    SG-PGM@3 \small{(ours)} & \textbf{95.41} & \textbf{95.01} & \textbf{95.21} \\ \hline
    \end{tabular}
    }
    \caption{\textbf{Overlap check for point cloud registration.}}
    \label{tab:sgm_overlap_more}\vspace{-2mm}
\end{table}

\section{Qualitative Results}
\label{sec:vis}
\begin{figure*}[htb]
     \centering
     \begin{subfigure}[b]{0.75\linewidth}
         \includegraphics[width=\textwidth]{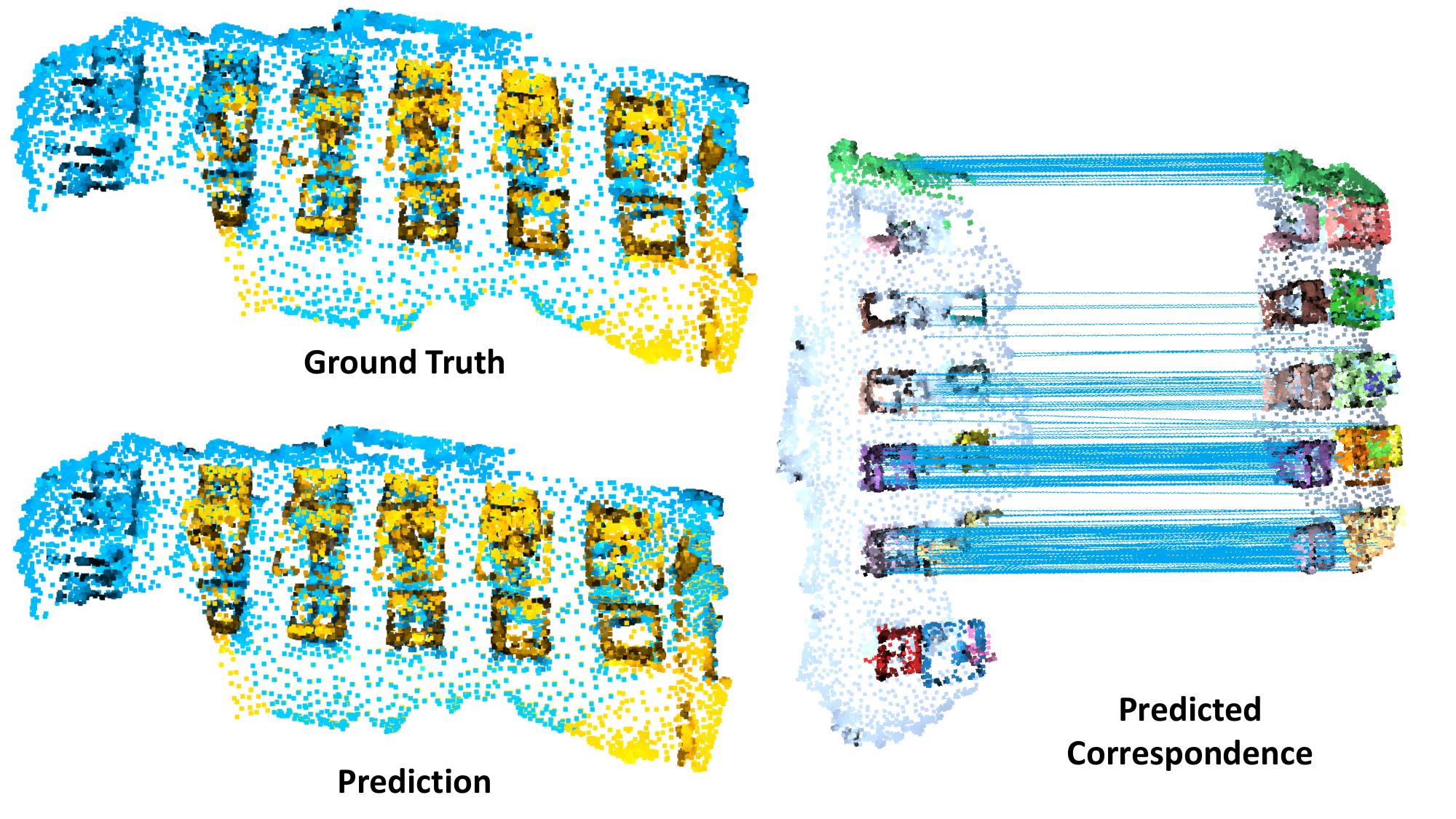}
     \end{subfigure}
     \\
     \begin{subfigure}[b]{0.75\linewidth}
         \includegraphics[width=\textwidth]{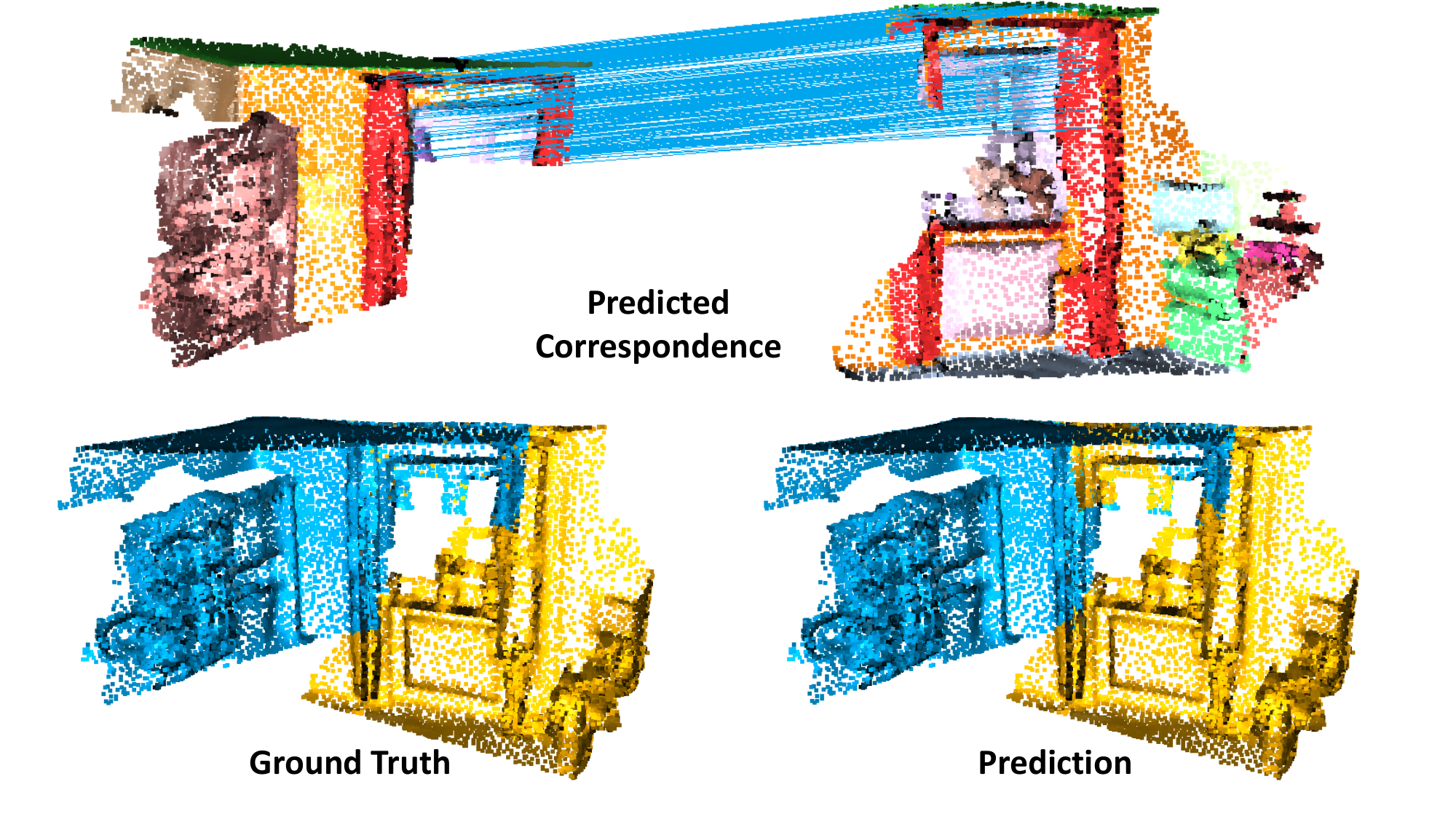}
     \end{subfigure}
     \\
     \begin{subfigure}[b]{0.75\linewidth}
         \includegraphics[width=\textwidth]{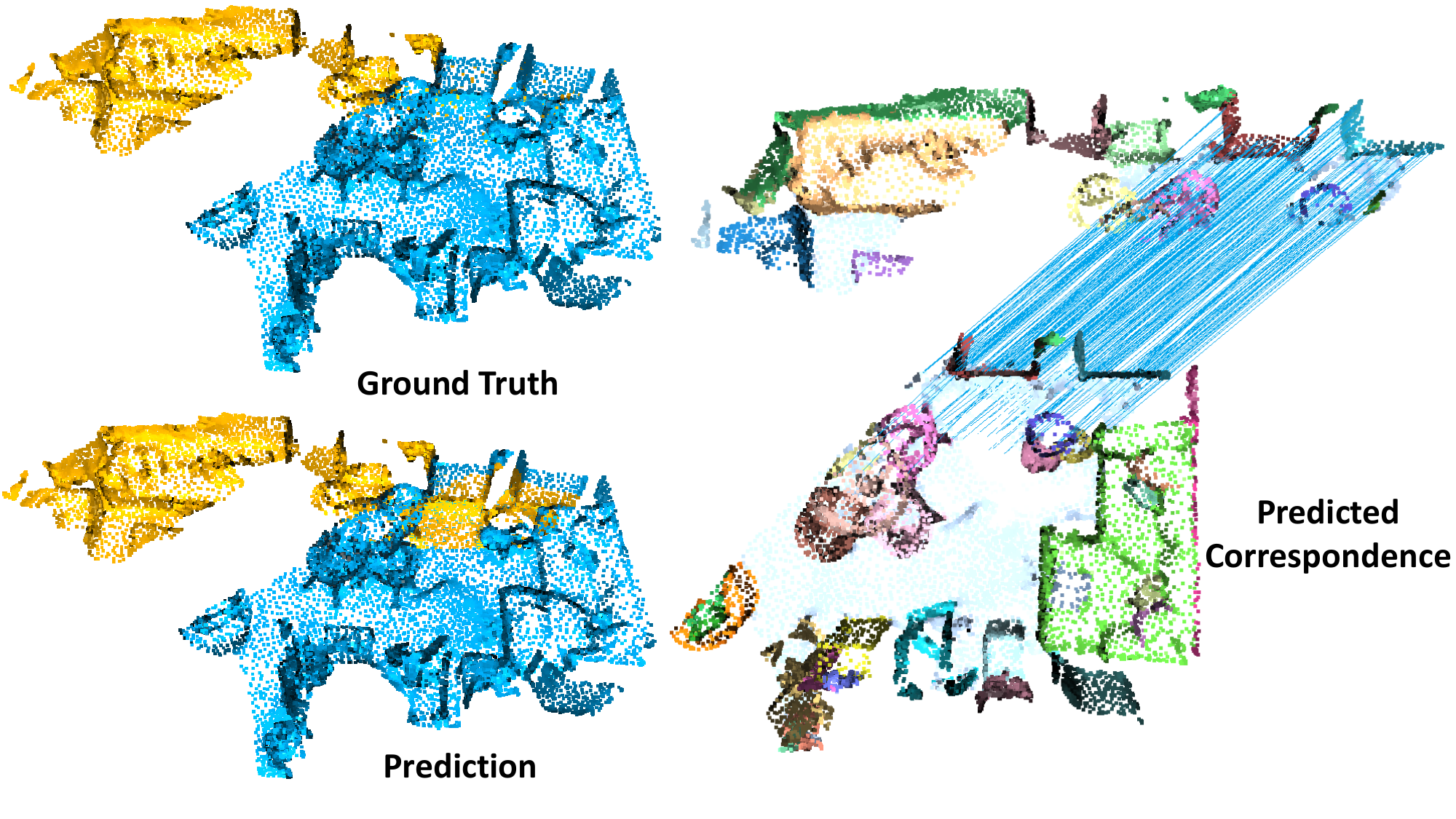}
     \end{subfigure}
     \caption{\textbf{Qualitative Results on Point Cloud Registration} of our proposed method.} 
    \label{fig:regis_vis}
\end{figure*}

\begin{figure*}[htb]
     \centering
     \begin{subfigure}[b]{0.75\linewidth}
         \includegraphics[width=\textwidth]{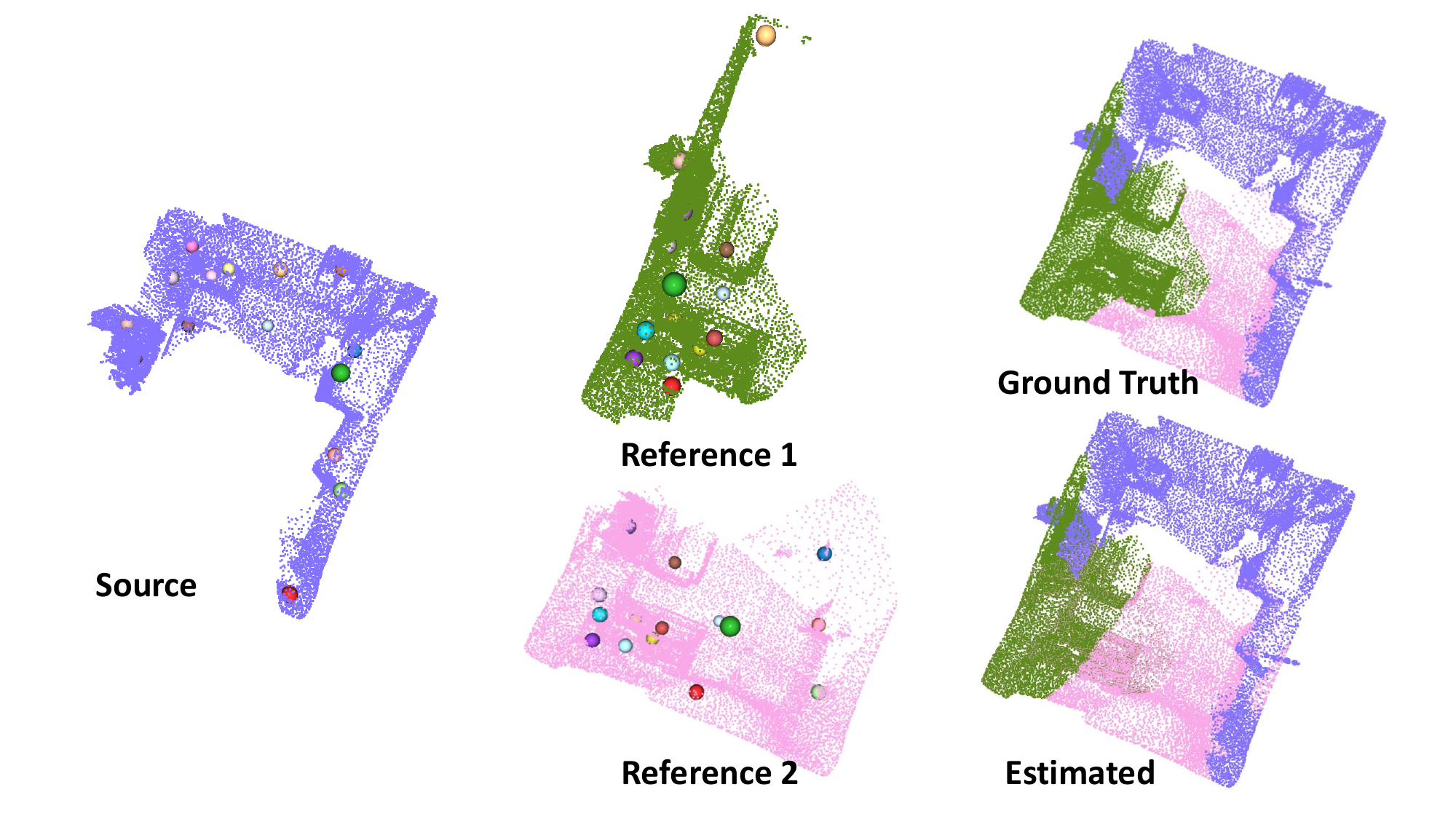}
     \end{subfigure}
     \begin{subfigure}[b]{0.75\linewidth}
         \includegraphics[width=\textwidth]{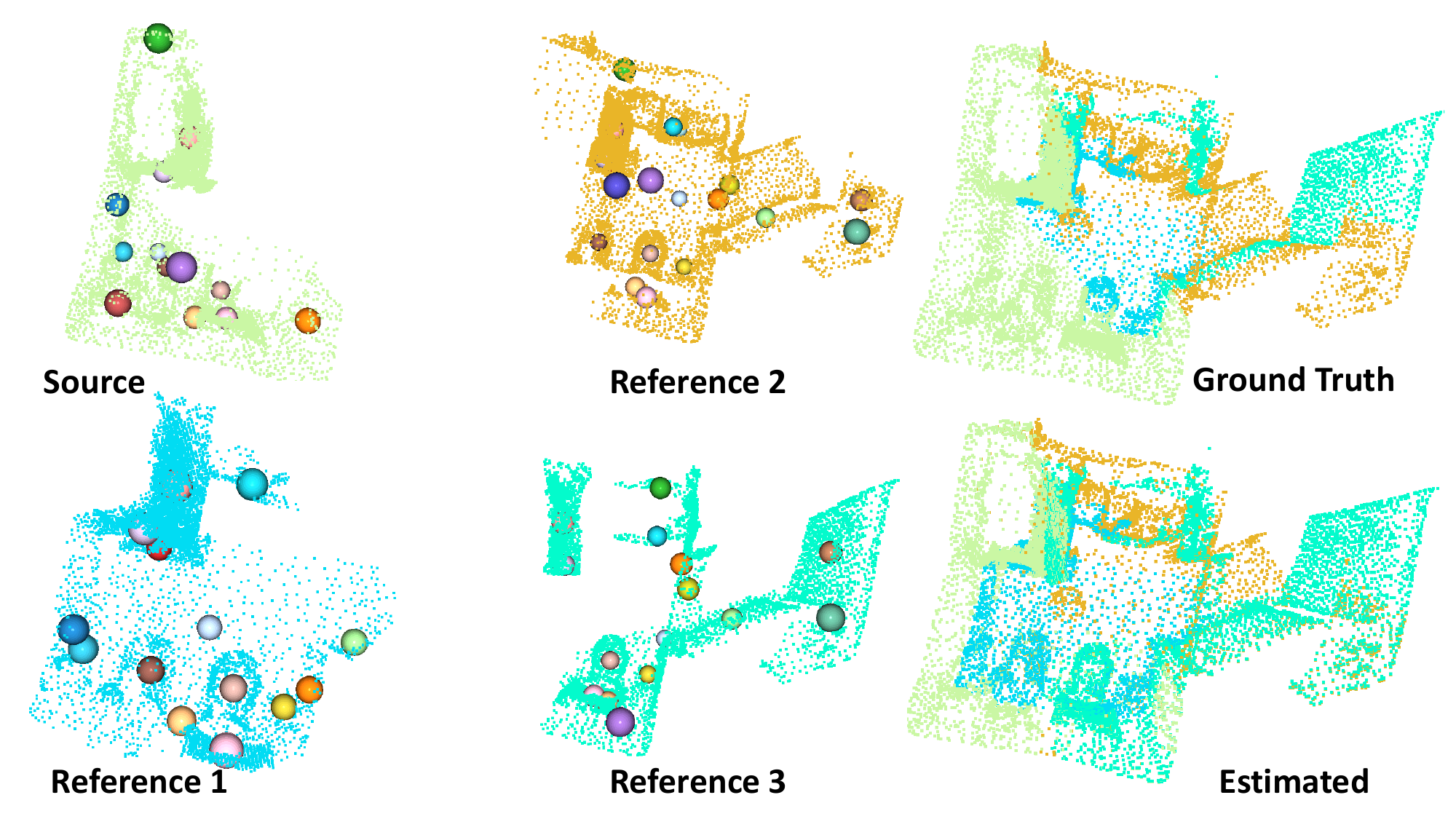}
     \end{subfigure}
     \begin{subfigure}[b]{0.75\linewidth}
         \includegraphics[width=\textwidth]{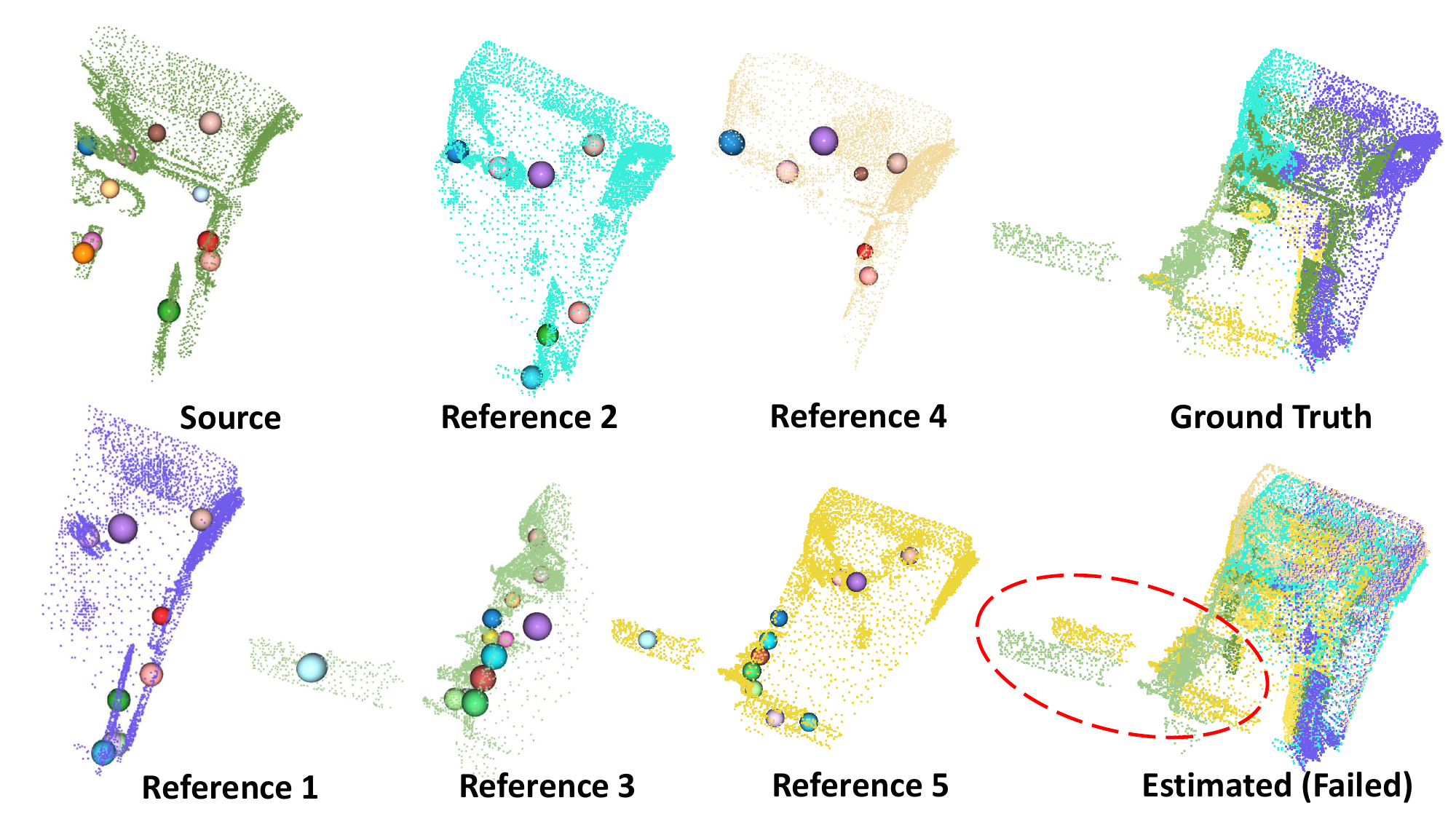}
     \end{subfigure}
     \caption{\textbf{Qualitative Results on Point Cloud Mosaicking} of our proposed method. Object nodes are visualized as 3D spheres.} 
    \label{fig:mosaick_vis}
\end{figure*}

Here we provide some qualitative results by combining our method and GeoTransformer~\cite{qin2022geometric} for point cloud registration in Figure~\ref{fig:regis_vis} and for point cloud mosaicking in Figure~\ref{fig:mosaick_vis}. 

\end{document}